\documentclass[preprint,11pt,3p,authoryear]{elsarticle}
\usepackage{hyperref}
\usepackage{lineno}
\usepackage{graphicx}
\usepackage{grffile}
\usepackage{amsmath,amssymb,amstext,amsthm}
\usepackage{natbib}
\usepackage{rotating}
\usepackage{adjustbox}
\usepackage[english,lined,boxed,ruled,linesnumbered,vlined]{algorithm2e}
\usepackage{titlesec}
\usepackage{multirow}
\usepackage{subcaption}
\usepackage{fancyhdr,mathtools,url}
\usepackage{wrapfig,float}
\usepackage{setspace}
\usepackage{booktabs,pdflscape,placeins,lastpage}
\usepackage{colortbl}

\definecolor{SolutionColor}{rgb}{0.9,0.9,1.0}
\usepackage{xcolor, soul}
\sethlcolor{SolutionColor}
\usepackage{caption}
\usepackage{longtable}
\usepackage{lscape}

\allowdisplaybreaks

\journal{}

\begin{document}

\begin{frontmatter}


\author[ufop]{J\'{u}lia C.~Freitas\corref{cor1}}
\ead{julia.freitas@aluno.ufop.edu.br}
\author[ufop]{Puca Huachi V.~Penna}
\ead{puca@ufop.edu.br}
\author[ufop]{T\'{u}lio A. M. Toffolo}
\ead{tulio@toffolo.com.br}
\cortext[cor1]{Corresponding author}
\address[ufop]{Federal University of Ouro Preto, Departament of Computing -- Brazil }

\title{Exact and Heuristic Approaches to Drone Delivery Problems} 



\address{}

\begin{abstract}
    The Flying Sidekick Traveling Salesman Problem (FSTSP) considers a delivery system composed by a truck and a drone. The drone launches from the truck with a single package to deliver to a customer. Each drone must return to the truck to recharge batteries, pick up another package, and launch again to a new customer location. 
This work proposes a novel Mixed Integer Programming (MIP) formulation and a heuristic approach to address the problem. 
The proposed MIP formulation yields better linear relaxation bounds than previously proposed formulations for all instances, and was capable of optimally solving several unsolved instances from the literature. 
A hybrid heuristic based on the General Variable Neighborhood Search metaheuristic combining Tabu Search concepts is employed to obtain high-quality solutions for large-size instances. The efficiency of the algorithm was evaluated on 1415 benchmark instances from the literature, and over 80\% of the best known solutions were improved.
 
\end{abstract}


\end{frontmatter}



\section{Introduction} \label{sec:Introduction}

Unmanned Aerial Vehicles (UAVs), generally known as drones, have recently shown a lot of potential in a wide range of civil purposes.
In the last years, with the quickly evolving of drone technology,  these vehicles and their applications have been attracting the attention of both the academic literature and large transportation companies \citep{Drone_Applications2018}.

Drones are employed within many applications in different fields. 
They are no longer limited to military use, and different businesses are now investing in these devices for faster and more responsive customer service. 
One of the most important purposes for these unmanned vehicles lies in disaster management, rescue operations and health-care. 
Drones can operate in dangerous environments that are inaccessible to humans. 
Another potential use concerns law enforcement, given these devices have the innate ability to hover around locations without drawing much attention from people. 
Thus, surveillance and public safety are also prominent applications of drones.
Drone delivery is another important and innovative application of drones, one which has become particularly relevant with the social distancing requirements due to the Covid-19 pandemic \citep{Guillot2020, Douglas2020}. 
The potential of drone delivery is vast both concerning operational cost and customer service efficiency.
However, these aerial vehicles are not a replacement for the traditional delivery trucks, due to their low payload capacity and short flying range. 
Drones can, however, be a useful complementary feature to the delivery process since they are not limited by road networks.
The applications of unmanned vehicles spread far beyond the preceding examples.
For a complete survey on drone applications and insights into emerging modeling approaches, the reader is referred to \cite{Otto2018}.

\begin{figure*}[!htp]
    \centering
    \begin{subfigure}[t]{0.5\textwidth}
        \centering
        \includegraphics[width=0.75\textwidth]{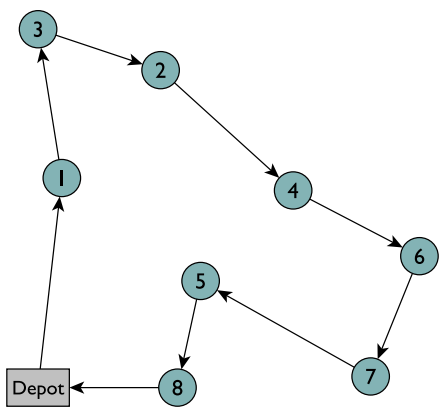}
        \caption{TSP -- the truck visits all customers.}
        \label{fig:tsp}
    \end{subfigure}%
    ~ 
    \begin{subfigure}[t]{0.5\textwidth}
        \centering
        \includegraphics[width=0.75\textwidth]{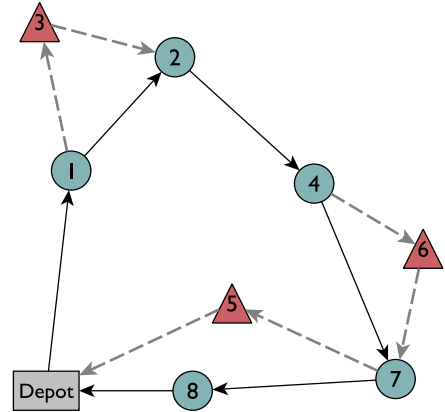}
        \caption{FSTSP approach, with assignment of customers to drone or truck.}
        \label{fig:fstsp}
    \end{subfigure}
    \caption{Customer deliveries are made by either a traditional delivery truck or via drone.}
    \label{fig:problem}
\end{figure*}

This paper focuses on a groundbreaking delivery modality including drones considered within the Flying Sidekick Traveling Salesman Problem (FSTSP), first introduced by \cite{Murray2015}. 
The FSTSP is a generalization of the Traveling Salesman Problem (TSP), one of the most well known and studied problems in operational research. 
While the classical TSP consists in finding an optimal route for one vehicle to deliver goods to a set of customers (Figure~\ref{fig:tsp}), the FSTSP consists in finding optimal routes for both a drone and one traditional delivery vehicle to perform goods distribution (Figure~\ref{fig:fstsp}). 
In Figure~\ref{fig:problem}, circular nodes indicate customers served by the truck while triangular nodes represent those served by the drone. 
Dashed and continuous lines indicate a drone trip and a truck tour, respectively. 
The delivery truck departs from the depot carrying a drone and all customer parcels. 
As the driver performs deliveries, the drone is launched from the truck, carrying the parcel for an individual customer. 
While the drone is on a trip, no intervention from the delivery driver is required. 
After delivering the parcel, the drone returns to the truck in a new customer location. 
One of the advantages associated with the truck-drone delivery system is the efficiency enhancing, as the drone capability to reach to more customers increases once it is launched from the truck closer to the customer delivery location.

The contribution of this paper is twofold.
A Mixed Integer Programming (MIP) formulation is proposed for the problem, one capable of solving several unsolved instances from the literature, and a heuristic to find high-quality solutions is developed to approach larger instances. 
Both the MIP model and the proposed heuristic were designed to tackle the two variants of drone delivery problems addressed by \cite{Murray2015} and \cite{Agatz2018}. 
The heuristic improved the best-known solution of 1138 instances from the literature (80.4\% of the total).
These results show the improvements that may occur in the total delivery time when using specific neighborhood structures of the problem within the local search.
The results also indicate that combining truck and drone for last-mile parcel delivery results in huge improvements in delivery time over truck tours for the instances considered.

In summary, this work presents both exact and heuristic approaches for the FSTSP, resulting in improvements over the best known bounds and solutions for a vast number of problem instances from the literature.

This manuscript is organized as follows. 
Section \ref{sec:Problem-Statement} details the problem. 
Section \ref{sec:Related_Literature} presents a few drone applications and a literature overview concerning the FSTSP. 
A novel MIP formulation for the problem is proposed in Section \ref{sec:Formulation}.
The heuristic algorithm developed is introduced and described in Section \ref{sec:Proposed-Methodology}. 
Section \ref{sec:Experimental-Analysis} presents the computational experiments and, finally, Section \ref{sec:Conclusions} summarizes the conclusions. 


\section{Problem description} \label{sec:Problem-Statement}

The Flying Sidekick Traveling Salesman Problem (FSTPS) presented by \cite{Murray2015} can be described as follows. 
Let $G= (V,A)$ be an undirected graph with $|V| = n+1$ nodes. 
Node $v_0 \in V$ represents the depot, where drone and truck must depart from and return to exactly once. 
The two vehicles (drone and truck) may depart and return from the depot either in tandem or independently. 
While traveling in tandem, the drone is transported by the truck to conserve battery. 
Every other node $v \in V \backslash \{v_0\}$ represents a customer that must be visited exactly once. 
Some customers must be served by the truck due to the travel time being superior to drone endurance, while others may be server by either the truck or the drone.

The vehicles do not necessarily follow the same distance metric. 
The truck is limited to the road network, while the drone can use a different network to travel between customers.

During the delivery, the drone may make multiple trips, each composed of three locations. 
The trip begins at the \textit{launch} node, which can be either the depot or a customer location. 
Before launch, the setup time $s^L$ is required for the drone to have its battery changed and the parcel loaded.
The second node in a trip is the \textit{delivery} node, which represents a customer serviced by the drone. 
The final or \textit{return} node is the location where the drone is collected.
This node may be either the depot or a location visited by the truck.
An additional setup time $s^R$ is required for the \textit{return} node since the drone must be recovered. 
We consider that whenever the drone is launched, the package will be successfully delivered and the drone will return to the truck or the depot without any issue, and within the drone's flight endurance limit. 
If the drone trip ends at the depot, the drone cannot be re-launched. 

The drone can only be recovered at a different position than the launch location in the truck route. Therefore, the drone must not be launched multiple times from the same location.
The drone may visit only one customer per trip. The truck may serve other customers in between launch and return locations, while the drone is serving a customer. Moreover, either of the two vehicles may have to wait for the other one depending on their arrival times at the return node. The waiting time must not exceed the drone's battery power.
Regarding vehicle capacity, the drone can carry and deliver only one parcel per trip, while no limit is imposed on the truck's capacity.

The FSTSP objective is to minimize the time required to complete all deliveries and return both vehicles to the depot.


\section{Related Literature} \label{sec:Related_Literature}
This section provides a literature review of the potential benefit of pilotless technology. 
This review first presents a few promising drone applications. 
Next, it aims at distinguishing the different approaches and delivery problem with drones.

Delivery applications have recently received considerable media attention, mainly because of the prospect of door-to-door express deliveries at low-cost. With this new concept, delivery times and costs could be significantly reduced. 
Because of the overwhelming amount of announcement in drone research by the industry, the literature has been growing significantly. 
Hereafter, we present some approaches that address the FSTSP and similar problems.


\cite{Murray2015} proposed a mixed integer linear programming formulation for two delivery-by-drone problems in which the delivery is performed by truck and drone. 
In the Parallel Drone Scheduling Traveling Salesman Problem (PDSTSP) the drone attends customers within flight range of the depot while the truck attends the remaining. 
The operations occur independently, i.e., while the truck follows a TSP route the drones fly to the customers and back to the depot multiple times. In the other problem, the truck and drone work collaboratively (Flying Sidekick Traveling Salesman Problem - FSTSP). The authors  implemented a simple and effective heuristic approach for both problems.
A similar work is the one by \cite{Boumanetal2017}. They addressed an exact solution approach for the TSP-D based on dynamic programming that can solve problems up to 16 customers. This number of customers is larger than the mathematical programming approaches presented in the literature thus far.
\cite{Ponza2016} based his dissertation on the FSTSP proposed by \cite{Murray2015}. He proposed a slightly different mathematical formulation to the problem and presented an analysis of several heuristics that could be used to resolve the problem. Moreover, he introduced a new set of instances to the literature.

\cite{DellAmico2020} approached the PDTSP proposing a MILP model and several matheuristics. The authors experiment with the algorithms on the benchmark instances introduced by \cite{MbiadouSaleu2018} and \cite{Murray2015}. The computational study validates that the proposed algorithms produce competitive results in terms of both efficiency and effectiveness mainly on small and medium-size instances. 

\cite{Wang2017} study the Vehicle Routing Problem with Drones (VRPD) from a worst-case point of view. The paper describes several theorems formulation for the vehicle routing problem with drones and represents bounds on maximal savings to the companies. \cite{Poikonen2017} expand the description of the theorems comparing different drone configurations in the delivery process to determine the maximum benefit. For example, the trade-off between speed and the number of drones, i.e., they compare what is better, a more substantial number of slower drones or a smaller number of faster drones. 

While the works above aim to minimize the time required to complete the tour, in \cite{Ha2018} the objective is to minimize the total operational cost of a drone-truck delivery system. They proposed a MILP formulation to the problem and a heuristic called TPS-LS, both inspired on the work of \cite{Murray2015}. Furthermore, a GRASP heuristic was presented. The results showed that GRASP provides better solution quality while TPS-LS deliver a lower solution quality, yet very quickly.

Another approach for the drone-truck system is the one presented in \cite{Jeong2019}. They considered two practical issues to evaluate the problem, the effect of parcel weight on drone energy consumption and restricted flying areas. Their sensitivity analysis
shows that the increase in package weight and no-fly zones reduce the efficient use of drones as it limits their flight range, especially when the two factors are combined.

The FSTSP proposed by \cite{Gonzalezetal2020} allows the truck to wait for the drone where it was launched. The drone also can perform multiple visits per launch. The authors considered drone energy, i.e., the battery is changed between drone trips and it is considered fully charged after the swap. An iterative greedy search heuristic combined with simulated annealing was proposed.

\cite{FreitasPenna2020} introduces new instances based on the TSPLIB and compares the HGVNS (General Variable Neighborhood Search) result with instances found in the literature \citep{Agatz2018, Ponza2016}. In this work, we complement the heuristics by using a list, based on the Tabu Search, to avoid cycling in the neighborhoods. Here, we also propose a MILP 
to solve the FSTSP in the instances proposed in \citep{Murray2015, Ponza2016}.


The multiple flying sidekicks traveling salesman problem (mFSTSP) introduced by \cite{Murray2020} considers a delivery truck operating in coordination with a fleet of drones.
The drones are launched from the truck to deliver a single package, then return to the truck where it can be loaded again.
They employed a three-phased (I. initial truck assignments, II.create drone routes, III.combining phase I and II) heuristic solution to approach the problem. The heuristic result analysis revealed that drones with high-speed and long-range offer greater benefits in larger geographic regions, where customers are distributed over a larger area.

\cite{Ferrandezetal-2016} introduced a truck-drone delivery system where multiples drones travel per truck. They investigated the time and energy associated with a truck-drone delivery network compared to a standalone truck or drone delivery. Besides, they proposed a k-means and a genetic algorithm to determine the optimal number of launch locations and drones per truck.
Following the multiple drones per truck approach, \cite{Karak2019} presents a mathematical formulation and solution methodology for the hybrid vehicle-drone routing problem (HVDRP) for pick-up and delivery services. The problem is formulated as a mixed-integer program, which minimizes the vehicle and drone routing cost to serve all customers. The formulation captures the vehicle-drone routing interactions during the drone dispatching and collection processes and accounts for drone operation constraints related to flight range and load carrying capacity limitations. A novel solution methodology is developed which extends the classic Clarke and Wright algorithm (HCWH) to solve the HVDRP. The performance of the developed heuristic is benchmarked against two other heuristics, namely, the vehicle-driven routing heuristic (VDH) and the drone-driven routing heuristic (DDH). A set of experiments are conducted to evaluate the performance of the developed heuristics. While the VDH and the
DDH focus on optimizing the cost of one mode only, the HCWH is shown to outperform these two heuristics in terms of minimizing
the cost of the entire multi-modal network. The network operation cost is shown to be minimum when the used drones are balanced in terms of their flight range and load carrying capacity.

\cite{Schermer2019, Wang2017, Poikonen2017, Ulmer2017, DiPugliaPugliese2017} tackled the Vehicle Routing Problem with Drones (VRPD). In \cite{Schermer2019}, they formulated the VRPD as a Mixed Integer Linear Program (MILP), and introduced several sets of valid inequalities aiming to improve the performance of solvers. To address large instances, they presented a matheuristic approach that exploits the problem structure of the VRPD.
The authors proposed the Drone Assignment and Scheduling Problem (DASP) defined as minimization problem that given an existing routing of trucks, looks for an optimal assignment and schedule of drones such that the makespan is minimized.


Contrasting with the others approaches, \cite{Song2018} addressed a delivery problem where drones serve all customers. The authors described the Unmanned Aerial Vehicle Routing Problem (UAVRP) presenting new features to drone delivery problem: capacity and time window. Multiple drones, package weight impacting the vehicle battery life and customers demand to be satisfied are also considered. Service stations are strategically positioned to respect the endurance of drones and minimize delivery time. In these stations vehicles can be recharged and reload; thus, the drones can serve customers persistently. Computation experiments evaluated a heuristic and a mixed integer linear programming (MILP) formulation. While the MILP formulation could not solve large-scale problems, the heuristic successfully derives optimal or near-optimal solutions for them in a short time. 
Concerning drone-only delivery \citep{Dorlingetal2018} presented in their work an energy consumption model for multirotor drones and provided a linear approximation for it. They proposed the Drone Delivery Problem (DDP) which seeks to minimize cost (MC-DDP) or delivery time (MT-DDP) while considering battery weight, payload weight, and drone reuse. They proposed a MILP implementation and a Simulated Annealing (SA) to solve practical scenarios with hundreds of locations. Comparing the approaches, the SA implementation consistently finds near-optimal solutions to problems with eight or fewer locations. The heuristic behavior in larger instances with 125 to 500 customers showed consistent results.
\cite{Arenzana2020} work presented a strategic framework to quantify the efficiency of hospital operations. They introduced a MILP to design drone delivery network for hospital deliveries. The problem minimizes drone travelling time, battery consumption levels, vehicle investment, and infrastructure costs. The trajectories designed between hospitals conform to the latest air traffic management regulations. They presented a case study based in London. The paper shows that drones present numerous advantages in comparison with traditional road transport. With operational costs averaging 30\% depending on drone model and operational
parameters, such as vehicle range and payload size, the drone-based model increases service reliability (lower variability in travel time) and overcomes initial investment.

Table \ref{table:comparison-related-work} summarises the related work mentioned in this section. Column \emph{Problem class} defines the problem approached in the paper. \#Drones and \#Trucks describe, respectively, the number of drones and trucks considered in the problem. Column Same L/R indicates the node where the drone is launched is necessarily equal from the node it returns.
Same net describes if the network the vehicles travel are the same.
Then, there are three columns describing drones: 1) whether energy consumption is evaluated (Endurance), 2) whether capacity is considered (Capacity), 3) whether drone can perform multiple visits (Drone multiple visits). Column Sync indicates if trucks and drones perform a coordinated operation. Finally, the last column reports the works that studied the problem.

\begin{landscape}
\begin{table}[H]
  \small
  \centering
  \caption{Summary of the main features of FSTSP contributions in the literature.}
  \label{table:comparison-related-work}%
  \begin{adjustbox}{}
    \begin{tabular}{lccccccccr}
    \toprule
 \multicolumn{1}{c}{Problem} &  & & \multicolumn{1}{c}{Same} & \multicolumn{1}{c}{Same} &  &  & \multicolumn{1}{c}{Drone}  &  & \multicolumn{1}{c}{Related} \\
\multicolumn{1}{c}{class} & \multicolumn{1}{c}{\#Drones} & \multicolumn{1}{c}{\#Trucks} & \multicolumn{1}{c}{L/R}  & \multicolumn{1}{c}{net} & \multicolumn{1}{c}{Endurance} & \multicolumn{1}{c}{Capacity} & \multicolumn{1}{c}{multiple}  & \multicolumn{1}{c}{Sync} & \multicolumn{1}{c}{Work} \\
   & & & \multicolumn{1}{c}{} & \multicolumn{1}{c}{} &  &  & \multicolumn{1}{c}{visits} &  &  \\
          \midrule
          
\multirow{4}{*}{FSTSP} & \multirow{4}{*}{1} & \multirow{4}{*}{1} & \multirow{4}{*}{$\times$} & \multirow{4}{*}{$\times$} & \multirow{4}{*}{$\times$} & \multirow{4}{*}{$\times$} & \multirow{4}{*}{$\times$} & \multirow{4}{*}{\checkmark}  & \cite{Murray2015} \\
& & & & & & & & & \cite{Ponza2016} \\
& & & & & & & & & \cite{Ha2018} \\
& & & & & & & & & \cite{FreitasPenna2020} \\
\arrayrulecolor{lightgray}\midrule

FSTSP & 1 & 1 & $\times$ & $\times$ & $\times$ & $\times$ & \checkmark & \checkmark  & \cite{Gonzalezetal2020} \\
\arrayrulecolor{black}\midrule

\multirow{2}{*}{PDSTSP} & \multirow{2}{*}{n} & \multirow{2}{*}{1} & \multirow{2}{*}{$\times$} & \multirow{2}{*}{$\times$} & \multirow{2}{*}{$\times$} & \multirow{2}{*}{$\times$} & \multirow{2}{*}{$\times$} & \multirow{2}{*}{$\times$}  & \cite{Murray2015} \\
& & & & & & & & & \cite{DellAmico2020} \\
\midrule
\multirow{2}{*}{TSP-D} & \multirow{2}{*}{1} & \multirow{2}{*}{1} & \multirow{2}{*}{\checkmark} & \multirow{2}{*}{\checkmark} & \multirow{2}{*}{$\times$} & \multirow{2}{*}{$\times$} & \multirow{2}{*}{$\times$} & \multirow{2}{*}{\checkmark}  & \cite{Boumanetal2017} \\
& & & & & & & & & \cite{Agatz2018} \\
\midrule
\multirow{5}{*}{VRPD} & \multirow{5}{*}{n} & \multirow{5}{*}{m} & \multirow{5}{*}{$\times$} & \multirow{5}{*}{\checkmark} & \multirow{5}{*}{$\times$} & \multirow{5}{*}{$\times$} & \multirow{5}{*}{$\times$} & \multirow{5}{*}{\checkmark}  & \cite{Wang2017} \\
& & & & & & & & & \cite{Poikonen2017} \\
& & & & & & & & & \cite{Schermer2019} \\
& & & & & & & & & \cite{Ulmer2017} \\
& & & & & & & & & \cite{DiPugliaPugliese2017} \\
\midrule
VRPDR & n & m & \checkmark & $\times$ & \checkmark & \checkmark & $\times$ & \checkmark & \cite{Dayarian2020} \\
\arrayrulecolor{lightgray}\midrule\arrayrulecolor{black}
VRPDR & 1 & 1 & $\times$ & $\times$ & \checkmark & \checkmark & $\times$ & \checkmark & \cite{Dayarian2020} \\
\midrule
DDP & n & 0 & \checkmark & - & \checkmark & \checkmark & $\times$ &  -    & \cite{Dorlingetal2018} \\
\midrule
UAVRP & n & 0 & \checkmark & - & \checkmark & \checkmark & \checkmark &  -    & \cite{Song2018} \\
\midrule
FSTSP-ECNZ & 1 & 1 & \checkmark & $\times$ & \checkmark & \checkmark & $\times$ & \checkmark  & \cite{Jeong2019} \\
\midrule
mFSTSP & n & 1 & $\times$ & $\times$ & $\times$ & \checkmark & $\times$ & \checkmark  & \cite{Murray2020} \\
\midrule
HVDRP & n & 1 & \checkmark & \checkmark & $\times$ & \checkmark & \checkmark & \checkmark  & \cite{Karak2019}   \\
\midrule
STRPD & n & 1 & \checkmark & $\times$ & $\times$ & \checkmark & \checkmark & \checkmark  & \cite{MoshrefJavadi2020}   \\
\bottomrule
    \end{tabular}%
    \end{adjustbox}
\end{table}%
\end{landscape}


\newenvironment{vardescription}[1]{\setlength{\itemsep}{2pt}
    \begin{list}{}{\renewcommand{\makelabel}[1]{\hspace{0cm}\hfil##1  :}%
            \settowidth{\labelwidth}{\hspace{0cm}\textbf{#1} :}%
            \setlength{\leftmargin}{\labelwidth}\addtolength{\leftmargin}{\labelsep}}}%
        {\end{list}}

\section{Formulation} \label{sec:Formulation}

    This section proposes a compact Mixed Integer Programming (MIP) formulation for the FSTSP.
    As with the problem description in Section~\ref{sec:Problem-Statement}, a graph $G=(V,A)$ is considered to represent the problem.
    Table~\ref{notation} presents the notation employed throughout the formulation.
    Note in this table that $L$ is the set of possible \emph{moments} for the truck to visit a customer. 
    It is defined such that $L = \{0,\ldots,n\}$, where $n$ is the number of customers. 
    This set contains indices used to select the order in which the customers are visited. 
    The customers' visiting order is crucial to synchronize the truck and the drone.

    \begin{table}[H]
        \centering
        \caption{Sets and input data utilized within the formulation}
        \label{notation}
        \begin{tabular}{cll}
            \toprule
            $V$              & vertex set including the depot and the $n$ customers, $V=\{v_0,\ldots,v_n\}$ \\
            $V'$             & vertex set excluding the depot, $V' = V \backslash \{v_0\}$ \\
            $A$              & arc set \\
            $D$              & set of possible drone paths $(i,k,j)$ formed by two arcs, $(i,k)$ and $(k,j)$ \\ & that respects the drone's maximum endurance \\
            $L$              & set of possible moments for truck visits, $L=\{0,\ldots,n\}$ \\
            \midrule
            $e$              & drone flight endurance time \\
            $s^L$            & setup time for launching the drone \\
            $s^R$            & setup time for returning the drone \\
            $\tau_{i,j}$     & time required by the truck to traverse arc $(i,j)$ \\
            $\tau^D_{i,k,j}$ & time required by the drone to traverse arcs $(i,k)$ and $(k,j)$ \\
            $M$              & upper bound for the time required by the truck to visit all customers \\
            \bottomrule
        \end{tabular}
    \end{table}

    The formulation considers three variable sets:
    
    \begin{vardescription}{xxxx}
        \item[$t^{\ell}$] variable that defines the total travel time until moment $\ell$;
        \item[$x^{\ell}_{i,j}$] binary variable equal to 1 if the truck traverses arc $(i,j)$ at moment $\ell$, and 0 otherwise;
        \item[$y^{\ell,\ell'}_{i,k,j}$] binary variable equal to 1 if the drone traverses arcs $(i,k)$ and $(k,j)$, launching from vertex $i$ at moment $\ell$ and returning to the truck in vertex $j$ at moment $\ell'>\ell$, and 0 otherwise.
    \end{vardescription}

    At first sight, the number of variables may seem prohibitively significant.
    However, in practice, this number can be considerably reduced by filtering variable sets $x$ and $y$ to consider only feasible connections, meaning only $(i,j) \in A$ and $(i,k,j) \in D$ should be considered.
    Moreover, despite requiring more variables, the formulation here proposed is stronger than those proposed by \cite{Murray2015} and \cite{Ponza2016}, yielding better linear relaxation lower bounds for all instances considered (computational results are presented in Section \ref{sec:Experimental-Analysis}).

    The formulation is presented by Equations~\eqref{fo}--\eqref{bin}.
    To simplify the notation and reduce the constraints length, we assume $x^{\ell}_{i,j}=0$ for all $(i,j)\notin A$ and, analogously, $y^{\ell,\ell'}_{i,k,j}=0$ for all $(i,k,j)\notin D$ and all nonexistent moment pairs $(\ell,\ell')$ with $\ell \geq \ell'$.
    Note that such variables are not generated by our implementation, whose source code is available online\footnote{The formulation's implementation is available at \url{http://www.github.com/tuliotoffolo/fstsp}}.

    \begin{alignat}{4}
        min. \quad
        \label{fo}                                                                                             &
        \mathrlap{t^{n+1}} \\
        %
        s.t.\quad
        \label{depot}                                                                                          &
        \sum_{j\in V} x^{0}_{v_0,j} = \sum_{j\in V}\sum_{\ell\in L\backslash\{0\}} x^{\ell}_{j,v_0} = 1                                                                                  \\
        \label{single_moment}                                                                                  &
        \sum_{j\in V} \sum_{\ell\in L} x^\ell_{i,j} = \sum_{j\in V} \sum_{\ell\in L} x^\ell_{j,i} \leq 1       &   & \quad\forall i\in V                                             \\
        \label{flow}                                                                                           &
        \sum_{j\in V} x^{\ell-1}_{j,k} = \sum_{j\in V} x^{\ell}_{k,j}                                          &   & \quad
        \forall k\in V', \ell \in L\backslash\{0\}               \\
        \label{one_arc}                                                                                        &
        \sum_{(i,j)\in A} x^\ell_{i,j} \leq 1                                                                  &   & \quad \forall \ell\in L                                          \\
        \label{one_drone}                                                                                      &
        \sum_{(i,k,j)\in D} \sum_{l=0}^{\ell}\sum_{l'=\ell+1}^{n} y^{l,l'}_{i,k,j} \leq 1                      &   & \quad \forall \ell\in L                           \\
        \label{all_customers}                                                                                  &
        \sum_{j\in V}\sum_{\ell\in L} x^{\ell}_{k,j}
        + \sum_{i\in V}\sum_{j\in V}\sum_{\ell\in L}\sum_{\ell'\in L} y^{\ell,\ell'}_{i,k,j} = 1               &   & \quad \forall k\in V'                                            \\
        \label{drone_launch}                                                                                   &
        \sum_{k\in V'}\sum_{j\in V}\sum_{\ell'\in L} y^{\ell,\ell'}_{i,k,j} \leq \sum_{j\in V}x^{\ell}_{i,j}   &   & \quad \forall i\in V, \ell \in L                                 \\
        \label{drone_return}                                                                                   &
        \sum_{i\in V}\sum_{k\in V'}\sum_{\ell\in L} y^{\ell,\ell'}_{i,k,j} \leq \sum_{i\in V}x^{\ell'-1}_{i,j} &   & \quad \forall j\in V, \ell' \in L                                \\
        \label{endurance}                                                                                      &
        t^{\ell'} - t^{\ell} \leq e
        + M \left(1 - \hspace{-0.2cm}\sum_{(i,k,j)\in D}\hspace{-0.2cm} y^{\ell,\ell'}_{i,k,j}\right)           &   & \quad
        \forall \ell\in L \backslash\{0\}, \ell'\in L: \ell'>\ell  \\
        \label{time_truck_only}                                                                                &
        \mathrlap{t^{\ell} \geq t^{\ell-1} + \sum_{(i,j)\in A} \tau_{i,j} x^{\ell'-1}_{i,j} + \sum_{\substack{(i,k,j)\in D,\\\ell>1}} \sum_{l'=\ell}^{n}  s^L y^{\ell-1,l'}_{i,k,j} + \sum_{(i,k,j)\in D} \sum_{l=1}^{\ell-1} s^R y^{l,\ell}_{i,k,j}} \nonumber \\
                                    & &   & \quad
                                    \forall \ell\in L\backslash\{0\}\cup\{n+1\}                \\
        \label{time_with_drones}                                                                               &
        t^{\ell'} \geq t^{\ell} + \sum_{(i,k,j)\in D} \left( s^L + \tau^D_{i,k,j} + s^R \right) y^{\ell,\ell'}_{i,k,j}                    &   & \quad
        \forall \ell'\in L\backslash\{0\}, \ell\in L: \ell < \ell' \\
        \label{t_positive}                                                                                     &
        t^{0} = 0                                                                                                                                                                    \\
        \label{bin-x}                                                                                          &
        x^{\ell}_{i,j} \in \{0,1\}                                                                             &   & \quad\forall (i,j) \in A, \ell \in L \\
        \label{bin}                                                                                            &
        y^{\ell,\ell'}_{i,k,j}\in \{0,1\}                                                                      &   & \quad\forall (i,j,k)\in D, \ell\in L, \ell'\in L : \ell'>\ell
    \end{alignat}

    The objective function presented by Equation~\eqref{fo} minimizes the total time to visit all customers, given by the sum of the truck's traveling time and all required setup times to launch and collect the drone.
    Constraints \eqref{depot} ensure the truck leaves the depot at moment zero and returns to it at the tour's end.
    Constraints \eqref{single_moment} limit the number of truck visits to any customer to one.
    Constraints \eqref{flow} are flow preservation constraints which force the truck to leave a customer at the subsequent moment of its visit.
    Constraints \eqref{one_arc} limit the number of arcs traversed at each moment to at most one.
    Constraints \eqref{one_drone} prohibit launching the drone more than once in overlapping time windows (given by $l$ and $l'$) and therefore assert the drone is not launched when it is not with the truck.
    Note that $l$ and $l'$ are used to cover all time windows, including moment $\ell$.
    Constraints \eqref{all_customers} guarantee every customer is visited exactly once, either by the truck or by the drone.
    Constraints \eqref{drone_launch} and \eqref{drone_return} synchronize the truck's position with the drone's launch and return, respectively.
    Constraints \eqref{endurance} certify the drone's endurance is respected.
    Note that these constraints employ a `\emph{Big M}', which disables the constraint whenever the drone is not launched. 
    The value of M is set to an upper bound\footnote{The upper bound was obtained by a simple Nearest Neighbor constructive heuristic, and its value is available as part of \cite{Ponza2016} instances.} on the time at which both the drone and the truck return to the depot.
    Also, the truck's travel time is not considered for endurance when the drone launches from the depot (when $\ell=0$).
    Constraints \eqref{time_truck_only} update the travel time until moment $\ell$ considering the truck's route. 
    Eventual setup times $s^L$ and $s^R$ of launching and returning the drone, respectively, are taken into account.
    Similarly, Constraints \eqref{time_with_drones} ensure the travel time until moment $\ell$ includes the time traveled by the drone and eventual setup times $s^L$ and $s^R$.
    Therefore, time $t^\ell$ of any moment $\ell>0$ considers the travel time of both truck and drone, including whichever is larger.
    Constraint \eqref{t_positive} sets the total travel time at the first moment to zero and, finally, Constraints \eqref{bin-x} and \eqref{bin} declare the binary nature of variables $x$ and $y$. 

\section{The Hybrid Heuristic} 
\label{sec:Proposed-Methodology}

The proposed heuristic algorithm, named Hybrid Tabu General Variable Neighborhood Search (HTGVNS), is a hybrid metaheuristic that combines the exact solution of a TSP solver and the exploration capabilities of systematical neighborhood changes. 
The HTGVNS employs the Route First Cluster Second approach of \cite{Beasley1983} in which first a TSP is solved and then clusters are created by assigning customers to the drone.

The following sections detail the heuristic approach. First Section \ref{subsection:neighborhoods} introduces the different neighborhood structures used within the algorithm. Then, Section \ref{subsection:HTGVNS_Algorithm} formally introduces the proposed heuristic. 

\subsection{Neighborhood structures}  \label{subsection:neighborhoods}
Several \emph{neighborhood structures} were developed to explore the search space of the FSTSP, based on different \emph{moves}.
When a move is applied on a reference solution $S$, a \emph{neighbor solution} $S'$ is obtained.
A neighborhood structure $N$ consists of all possible applications of a \emph{move}. 
Considering a reference solution $S$, this results in a set of neighbor solutions $N(S)$, which is hereinafter referred to as \emph{neighborhood}.

The neighborhood structures considered are based mostly on classical TSP moves, with some of them relying on specificities of the problem.
The solution space is visited applying the Best Improvement (BI) \citep{Hansen2017} approach, which exhaustively explores a neighborhood and returns the solution with the lowest objective value, i.e., the best neighbor.

\begin{figure*}[htb]
    \small
    \centering
    \begin{subfigure}[t]{0.5\textwidth}
        \centering
        \includegraphics[width=1\textwidth]{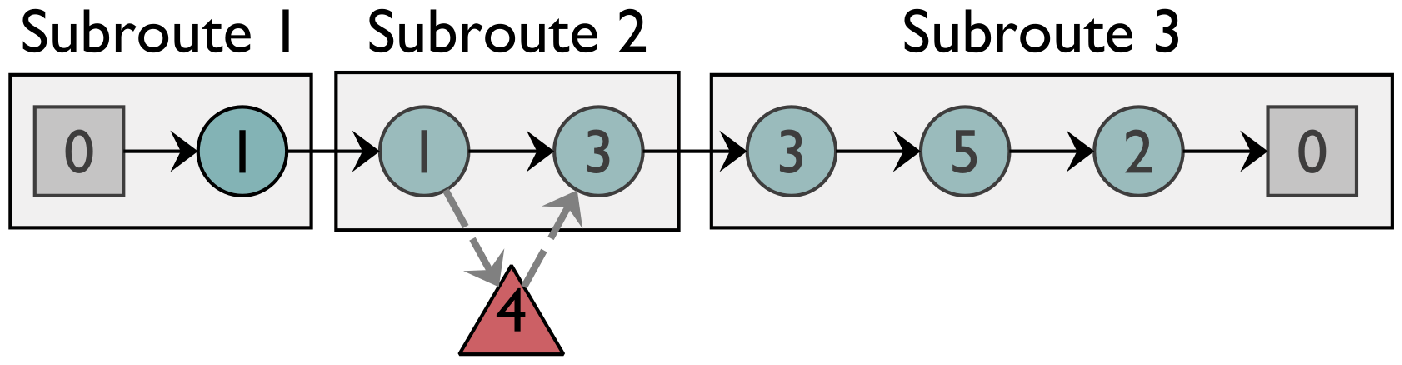}
        \caption{Example with one subroute.}
        \label{fig:subroute1}
    \end{subfigure}%
    ~ 
    \begin{subfigure}[t]{0.5\textwidth}
        \centering
        \includegraphics[width=1\textwidth]{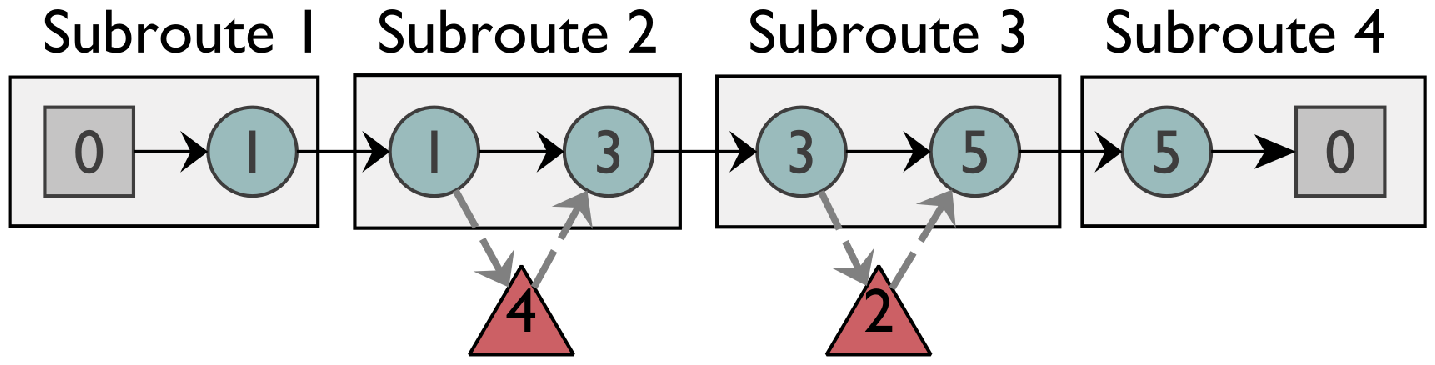}
        \caption{Example with two subroute.}
        \label{fig:subroute2}
    \end{subfigure}
    \caption{Customer deliveries are made by either a traditional delivery truck or via drone.}
    \label{fig:subroute}
\end{figure*}

Only feasible solutions are accepted within the proposed algorithm, meaning that infeasible neighbors are discarded. 
Note that to be feasible a route $(i,k,j)$ must respect the drone battery's life, i.e. $\tau^{D}_{i,k,j} \leq e$. 
Similarly, the drone battery power must endure until the truck arrival at the \textit{return} node.
Sub-routes are considered to avoid a drone launch before a return, as illustrated by Figure \ref{fig:subroute}, where each rectangle represents a sub-route. 
In Figure \ref{fig:subroute1}, drone customers can be assigned only to sub-routes 1 and 3. 
A launch must not occur in sub-route 2 since a drone trip is already associated with this sub-route. 
It is important to note that for every new drone trip, the truck route splits into one more sub-route. 
Therefore, new launches can only occur in sub-routes not associated with a drone. 
Figure \ref{fig:subroute2} describes the split of sub-route 3 as the result of a new drone trip.

The reader is directed to \cite{FreitasPenna2020} for more details on the neighborhood structures considered here, which are divided into two categories: intra-route and inter-route, detailed next. 

\subsubsection{Intra-route neighborhood structures}
The intra-route neighborhood structures considered are based on classical TSP moves. 

\begin{enumerate}[i.]
    \item \textsc{Swap(1,1)}: It swaps one customer in the solution with another one. 
    \item \textsc{Swap(2,1)}: It swaps two consecutive customers in the solution with another one. 
    \item \textsc{Swap(2,2)}: It swaps two consecutive customers in the solution with other two consecutive ones.
    
    \item \textsc{2-opt}:
    This move is the classical 2-opt move proposed by \cite{Croes1958}. Two edges are removed, and then, the two paths created are reconnected in the only possible way to keep a valid tour.
    
    \item \textsc{Reinsertion}:
    This move consists of removing a customer from its current position and reinserting it in another one.
    
    \item \textsc{Or-opt2}:
    This move consists of removing two consecutive customers from their current position and reinserting them in another one.
    
\end{enumerate}


\subsubsection{Inter-route Neighborhood structures}\label{sec:inter-route}
The inter-route neighborhood structures are based on moves which envision solution improvement by exploring the problem characteristics. 

\begin{enumerate}[i.]

\item \textsc{Shift(1,0)}:
This move consists of removing one truck customer and subsequently inserting it into the drone route. This move requires a cubic time, $\mathcal{O}(c^2c^\prime)$ where $c$ is the number of truck customers and $c^\prime$ is the number of eligible customers not currently assigned to the drone. This complexity time refers to three nested loops where the first defines the launch node $i$, subsequently, the delivery node $j$ and, lately, the return node $k$. This candidate trip must not exceed drone endurance.
The pair of nodes $i$ and $k$ are not necessarily adjacent, but $i$ must precede $k$.  
The selected combination of $\langle i,j,k \rangle$ is the one that presents the more considerable decreasing cost of the truck route within the removal of customer $j$.

\begin{figure}[th]
    \centering
    \begin{subfigure}[t]{0.5\textwidth}
        \centering
        \includegraphics[width=0.8\textwidth]{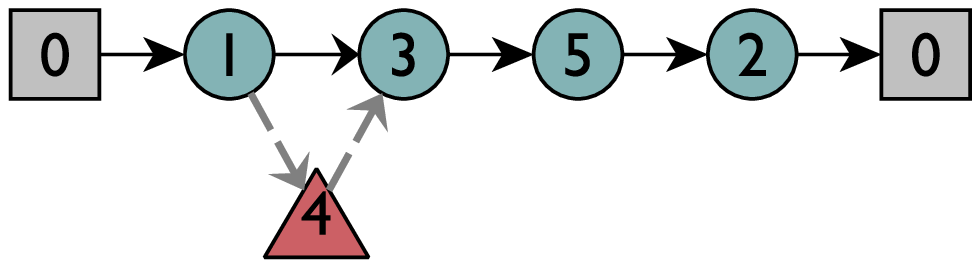}
        \caption{Original route.}
        \label{fig:move1_subroute1}
    \end{subfigure}%
    ~ 
    \begin{subfigure}[t]{0.5\textwidth}
        \centering
        \includegraphics[width=0.7\textwidth]{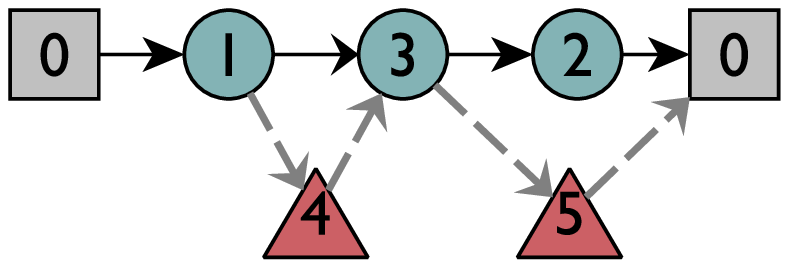}
        \caption{Insertion of a new drone trip.}
        \label{fig:move1_subroute2}
    \end{subfigure}
    \caption{Swap truck drone customer move.}
    \label{fig:move1}
\end{figure}

\item \textsc{Swap(1,1)}: This swap move consists of simply swapping two customers in the solution. One customer belongs to the truck and the other to the drone route. The complexity is $\mathcal{O}(c^\prime c)$ where $c^\prime$ and $c$ are the number of drone and truck customers, respectively. This complexity is due to the necessity of comparing all \textit{drone} nodes with every truck customer to find a combination at least as good as the previous or has a lower objective value. The swap is only performed if the drone endurance is not violated, i.e., the new drone trip can be completed before the drone runs out of battery power and the truck sub-route does not exceed the drone flight limit.

\begin{figure}[ht]
    \centering
    \begin{subfigure}[t]{0.5\textwidth}
        \centering
        \includegraphics[width=0.7\textwidth]{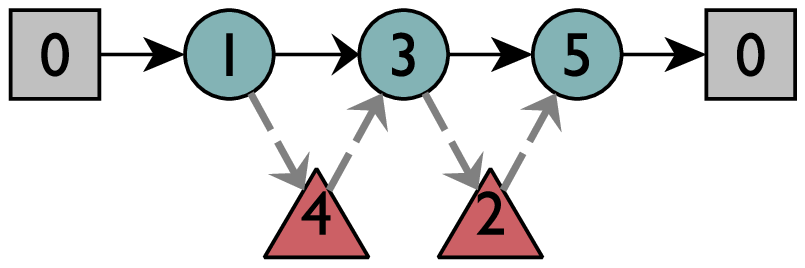}
        \caption{Original route.}
        \label{fig:move2_subroute1}
    \end{subfigure}%
    ~ 
    \begin{subfigure}[t]{0.5\textwidth}
        \centering
        \includegraphics[width=0.7\textwidth]{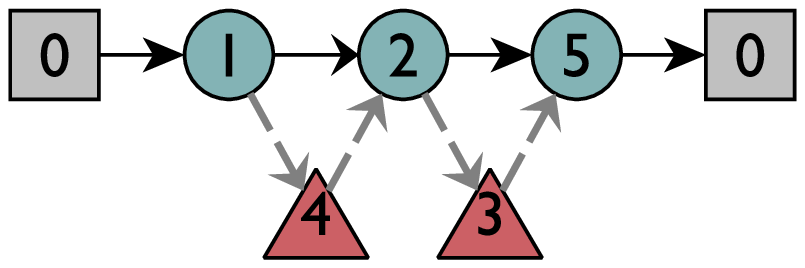}
        \caption{Swap of two customers.}
        \label{fig:move2_subroute2}
    \end{subfigure}
    \caption{Swap truck drone customer move.}
    \label{fig:move2}
\end{figure}

\item \textsc{Swap(0,1)}: This move is used to shake the current solution by turning a \textit{drone customer} into a \textit{truck customer}. A customer is removed from the drone route and inserted into the truck route in a position that generates the least impact on the solution quality. This move may increase the objective value, as it adds another customer to the truck route. However, if the truck waiting time is larger than the drone trip, a decrease in the objective value may occur.
This move requires combining \textit{drone customers}  with every position of the truck route to get the customer position that produces the smallest impact on the solution quality. Therefore, the time complexity of performing this move is $\mathcal{O}(c^\prime c)$ where $c^\prime$ and $c$ are the number of drone and truck customers, respectively. 

\begin{figure}[ht]
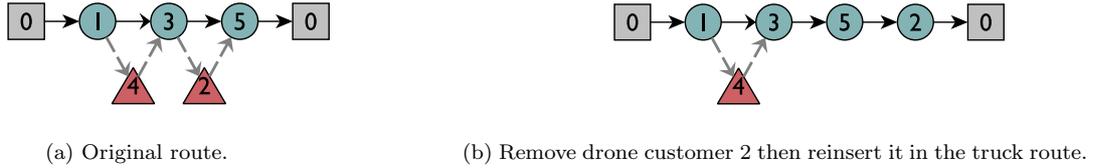

    \centering
    \begin{subfigure}[t]{0.5\textwidth}
        \centering
        \includegraphics[width=0.7\textwidth]{images/rota_original.eps}
        \caption{Original route.}
        \label{fig:move3_subroute1}
    \end{subfigure}%
    ~ 
    \begin{subfigure}[t]{0.5\textwidth}
        \centering
        \includegraphics[width=0.8\textwidth]{images/turn-drone-truck.eps}
        \caption{Remove drone customer 2 then reinsert it in the truck route.}
        \label{fig:move3_subroute2}
    \end{subfigure}
    \caption{Remove drone customer move.}
    \label{fig:move3}
\end{figure}

\end{enumerate}

\subsection{HTGVNS Algorithm}
\label{subsection:HTGVNS_Algorithm}

The proposed algorithm first builds an optimal TSP route by employing the exact approach from \cite{Concorde}. 
This results in a truck route with all customers, including eligible drone customers, which is used to build an initial solution to the FSTSP. 
This procedure, presented by Algorithm~\ref{algorithm:initialSolution}, requires two arguments: (\textit{i}) an initial TSP solution $S$ and (\textit{ii}) the eligible drone customers $C^\prime$. 
The algorithm iterates over the eligible drone customers until no improvement occurs.
For each customer, $j \in C^\prime$ the cost of removing it from the truck's route is computed (line~\ref{initialSolution:savings}). 
It is also verified for each subroute if it is associated with a drone trip (line~\ref{initialSolution:subroute}). 
For example, in Figure~\ref{fig:subroute1} \textit{Subroute}~1 is not associated with a drone trip while \textit{Subroute}~2 is. 
If the subroute is paired with a drone, an attempt is made to insert customer $j$ into the truck’s route between adjacent nodes $i$ and $k$ (line \ref{initialSolution:cost_truck}). 
Otherwise, the cost of serving customer $j$ by drone is computed (line \ref{initialSolution:cost_drone}).
Each pair of nodes $(i,k)$ such that $i$ precedes $k$ in a given subroute that is not currently connected with the drone is investigated. 
The goal is to compute the travel time associated with the drone trip launching from node $i$, visiting node $j$, and finally returning to the truck at node $k$.

\begin{algorithm}[!ht]
    \small
    \DontPrintSemicolon
    \caption{CreateInitialSolution}
    \label{algorithm:initialSolution}
  
    \SetKwInOut{Input}{\textbf{Input}}
    \SetKwInOut{Output}{\textbf{Output}}
    \Input{Eligible drone customer $C^\prime$ and optimal TSP solution $S$} \label{initialSolution:input}
      
      $truckSubRoutes \leftarrow$ \{$S$\}\;
        
    \textbf{stop} $\gets$ \emph{false}\;
    \While{\textbf{stop} $=$ false}{    
        \ForAll{j $\in$  $C^\prime$}
        {
            $savings \leftarrow$ calculate the savings of removing customer $j$ from the truck\textsc{\char13}s route\; \label{initialSolution:savings}
            \ForAll{$subroute \in$ $truckSubRoutes$}
            { \label{initialSolution:subroute}
                \eIf {$subroute$ is paired with drone} 
                {
                    $cost\_truck \leftarrow$ calculate the cost of relocating customer j in the truck\textsc{\char13}s route\;\label{initialSolution:cost_truck}
                }
                {  $cost\_drone \leftarrow$ calculate the cost of serving customer j with the drone\; \label{initialSolution:cost_drone}
                }
            }
        }
        $savings\_truck \leftarrow savings - cost\_truck$\;
        $savings\_drone \leftarrow savings - cost\_drone$\;
         \uIf{$savings\_truck < savings\_drone$}{
             update solution $S$ by moving customer $j$ to the drone\textsc{\char13}s route \;\label{initialSolution:update_truck}
            }
        \uElseIf{$savings\_truck > savings\_drone $}{
             update solution $S$ route by relocating customer $j$ in the truck\textsc{\char13}s route\;\label{initialSolution:update_drone}
            }
        \Else{
            {\textbf{\emph{stop}}} $\gets$ \emph{true}
        }
        }
        \Return S
\end{algorithm}

The proposed algorithm is a General Variable Neighborhood Search (GVNS), which consists of Variable Neighborhood Search (VNS) using Variable Neighborhood Descent (VND) as local search. 
\cite{Hansen2008} introduced the GVNS metaheuristic that relies upon a local search followed by perturbations to escape from local optima. 
Furthermore, to avoid cycling, it was adopted a tabu list likely in Tabu Search \citep{Glover1999} to forbid solutions that possess some attributes of recently explored solutions, especially between insertion and removal of drone customers. 
For example, in Figure~\ref{fig:subroute1} customer 4 was just added to the drone route as a \textit{drone customer}. Thus, it must not return to the truck route for the next $|TL|$ iterations, where $TL$ is the tabu list.


The Randomized Variable Neighborhood Descent (RVND) is an adaptation of the classic VND heuristic in which the neighborhood's selecting order is randomized. 
The VND is a heuristic described in \cite{Hansen2017} and further discussed by \cite{Souza2010} and \cite{Penna2013}, who showed that the randomized procedure often outperforms the deterministic approach. 
The RVND procedure developed for the FSTSP is presented in Algorithm~\ref{algorithm:RVND}.
The algorithm begins by filling the neighborhood structure list $\mathcal{N}$ (line \ref{RVNS:nlist}) and initializing solutions $S^*$ and $S'$ (line \ref{RVND:S}). 
Next, variable $k$ is initialized (line \ref{RVND:k_1}), representing the current number of neighborhoods analyzed.
The main loop (line \ref{RVND:mainloop}) first randomly selects a neighborhood structure (lines \ref{RVND:ifcomeco} -- \ref{RVND:iffim}). 
This structure can be chosen from two different sets, $\mathcal{N}$ or $\mathcal{N}$~$\backslash$~$\mathcal{N}_{TL}$. 
The first set is selected if the current solution $S^{\prime}$ is not in tabu list $TL$. Otherwise, the second set is selected. 
Afterwards, a new neighboring solution is generated (line \ref{RVND:newsolution}). 
If this solution is at least as good as the previous or has a lower objective value than the considered entry is accepted, the counter $k$ is reset to 1 and list $\mathcal{N}$ is shuffled (lines \ref{RVND:ifs} -- \ref{RVND:shuffle}). Otherwise, the value of $k$ is increased (lines \ref{RVND:elseifs} -- \ref{RVND:ife}). Once the maximum number of neighbors is reached, the best solution produced is returned (line \ref{RVND:end}).

\begin{algorithm}[!ht]
    \small
    \DontPrintSemicolon
    \caption{Randomized Variable Neighborhood Descent}
    \label{algorithm:RVND}
    \SetKwInOut{Input}{\textbf{Input}}
    \Input{ Solution $S$, tabu list $TL$, tabu neighborhood structures $\mathcal{N}_{TL}$}\label{RVND:input}
    Initialize and shuffle the list $\mathcal{N}$\ of neighborhood structures; \label{RVNS:nlist}\\
    $S^* \gets S^\prime \gets S$\; \label{RVND:S}
    $k \leftarrow$ 1 \; \label{RVND:k_1}
    \While{$k \le |\mathcal{N}|$}{\label{RVND:mainloop}

        \eIf{$S^{\prime}$ $\in$ $TL$} 
        { \label{RVND:ifcomeco}
            $N$ $\leftarrow$ random neighborhood structure from $\mathcal{N}$~$\backslash$~$\mathcal{N}_{TL}$\; 
             \label{RVND:if_lista_esta}
           }
           { \label{RVND:else_lista_esta}
            $N$ $\leftarrow$ neighborhood structure $\mathcal{N}^{(k)}$\; 
            \label{RVND:iffim}
           }
        $S^{\prime}$ $\leftarrow$ best neighbor solution in $N(S)$\; \label{RVND:newsolution}
        \eIf{$f(S^{\prime}) < f(S^*)$}{ \label{RVND:ifs}
            $S^*$ $\gets S^{\prime}$\; \label{RVND:update}
            $k \gets 1$\; \label{RVND:reset}
            shuffle $\mathcal{N}$ \label{RVND:shuffle}
        }
        {\label{RVND:elseifs} $k \gets k+1$ \;} \label{RVND:ife}
    } 
    \Return $S^*$\; \label{RVND:end}
\end{algorithm}


The combination of RVND and GVNS is presented in Algorithm \ref{algorithm:VNS}.
Three arguments are required: $(i)$ an initial solution $S$,  $(ii)$ the maximum number of interactions and $(iii)$ static tabu neighborhood list  $\mathcal{N}_{TL}$.


Algorithm \ref{algorithm:VNS} begins by initializing solution $S^*$, counter $k$, perturbation level $\rho$ and tabu list $TL$ (lines \ref{GVNS:initial-solution} -- \ref{GVNS:initial-tabu-list}). The main loop (line \ref{VNS:main-loop}) first generates a random neighbor solution $S$ from neighborhood $\mathcal{N}^{(k)}$ (line \ref{VNS:randomN}). If neighborhood $\mathcal{N}^{(k)}$ is in $\mathcal{N}_{TL}$, then solution $S$ is inserted in tabu list $TL$ (lines \ref{GVNS:if_lista_insere} -- \ref{GVNS:tl_insere}).
Afterwards, the RVND algorithm is called (line \ref{VNS:RVND}). If the solution produced by RVND is better than the current best solution, the best solution is updated (lines \ref{VNS:ifs} -- \ref{VNS:update1}). Note that counter $k$ and perturbation level $\rho$ are reset only if $S$ is an improving solution over $S^*$ (lines \ref{VNS:reset} -- \ref{VNS:reset_rho}). Otherwise, the value of $k$ is incremented (lines \ref{GVNS:else_k} -- \ref{GVNS:k}). 
Afterwards, the perturbation is executed (lines \ref{GVNS:shaking-begins} -- \ref{GVNS:shaking-ends}), which corresponds to applying neighborhood $\mathcal{N}^{(9)}$ (see subsection \ref{subsection:neighborhoods}) $\rho$ times to the current solution. Next, the perturbation level $\rho$ is increased (line \ref{GVNS:rho}) and the loop repeated. Variable $\rho$ is reset when it equals $\rho_{max}$.
Once the main loop reaches its stopping criterion, the best solution is returned (line \ref{VNS:return}).

\begin{algorithm}[!ht]
    \small
    \DontPrintSemicolon
    \caption{General Variable Neighborhood Search}
    \label{algorithm:VNS}
    
    \SetKwInOut{Input}{\textbf{Input}}
    \Input{ Initial solution $S$, maximum interaction $k_{max}$, maximum perturbation level $\rho_{max}$ and static tabu neighborhood list  $\mathcal{N}_{TL}$}\label{VNS:input}
    $S^*$ $\gets$ S\; \label{GVNS:initial-solution}
    $k \gets$ 1 \;
    $\rho \gets 0$\; \label{GVNS:pertubation-level}
    $TL \gets \emptyset$\; \label{GVNS:initial-tabu-list}
    \While{$k \le k_{max}$}
    { \label{VNS:main-loop}
        $S \gets$ random neighbor solution in $\mathcal{N}^{(k)}(S)$\;\label{VNS:randomN}
        \If{$\mathcal{N}^{(k)}$ $\in$ $\mathcal{N}_{TL}$\label{GVNS:if_lista_insere}}{
            insert $S$ in list $TL$ \;\label{GVNS:tl_insere}
        }
        $S \gets$ RVND($S$, $TL$, $\mathcal{N}_{TL}$) \; \label{VNS:RVND}
        \eIf{$f(S) < f(S^*)$}{ \label{VNS:ifs}
            $S^*$ $\gets S$\; \label{VNS:update1}
            $k \gets$ 1\;              \label{VNS:reset}
            $\rho$ $\gets 0$ \label{VNS:reset_rho}
        }
        {
            \label{GVNS:else_k} $k \gets k+1$\label{GVNS:k} 
        }
        \For{$i=0$ \KwTo $\rho$}{ \label{GVNS:shaking-begins}
           $S \gets$ random neighbor solution in $\mathcal{N}^{(9)}(S)$\; \label{GVNS:shaking-ends}
        }
         $\rho$ $\gets$ ($\rho$ $mod$ $\rho_{max}$ + 1)\;\label{GVNS:rho} 
    }
    \Return $S^*$\; \label{VNS:return}
\end{algorithm}


\section{Experimental Analysis} \label{sec:Experimental-Analysis}
In this section we present experimental results and analysis of the proposed exact and heuristic approaches. 
The MIP formulations were modeled using the Python-MIP package and solved using the commercial solver Gurobi 9.0 with the default configuration, while the heuristic algorithm was coded in C++ and compiled with G++ 5.3.1.
All experiments were executed on Intel Core i7 3.60GHz computers with 16GB of RAM running Ubuntu Linux 16.04.

First, Section \ref{sec:MIPModel} presents and analyses the results obtained by the proposed MIP formulation considering the instances from \cite{Murray2015} and \cite{Ponza2016}. Next, Section~\ref{sec:heuristic} discusses HTGVNS results for different benchmarks from the literature. 
Section~\ref{sec:ponza} considers the instances proposed by \cite{Ponza2016}, Section \ref{sec:agatz} considers those by \cite{Agatz2018} and, finally, the set introduced by \cite{FreitasPenna2020} is considered in Section \ref{sec:tsplib}. 

\subsection{Formulation Results} \label{sec:MIPModel}

The proposed formulation models the problem addressed by \cite{Ponza2016}. 
This problem considers, however, slightly different constraints than those considered by \cite{Murray2015}. There are two points of attention:
\begin{enumerate}
    \item in the problem described by \cite{Murray2015}, the truck's travel time between the drone's launch and return can be longer than the drone endurance; the drone can therefore run out of battery while waiting for the truck;
    \item \cite{Murray2015} do not consider the setup time for launching the drone as part of the drone's flying time, and it does not count for the total completion time or the battery's endurance, even when the drone leaves from the depot.
\end{enumerate}
Formulation \eqref{fo}--\eqref{bin} can be adapted to obtain results comparable with those by \cite{Murray2015} by removing Constraints \eqref{endurance} and altering Constraints \eqref{time_truck_only} and \eqref{time_with_drones}.

\begin{table}[b!]
    \small
    \centering
    \caption{Average number of variables and constraints per instance-set and endurance value}
    \label{VarConst}
    \begin{adjustbox}{max width=\textwidth}
    \begin{tabular}{lcccccccccc}
        \toprule
        \multirow{2}{*}{Instance set} & \multirow{2}{*}{\#nodes}  & \multirow{2}{*}{$e$} & \multicolumn{2}{c}{This work} && \multicolumn{2}{c}{\cite{Ponza2016}} && \multicolumn{2}{c}{\cite{Murray2015}} \\
        \cmidrule{4-5} \cmidrule{7-8} \cmidrule{10-11}
         &  &  & \#Vars & \#Constrs & & \#Vars & \#Constrs & & \#Vars & \#Constrs \\
        \midrule
        \cite{Ponza2016}  & 5 & 1,440 &  658 &  158 & &  116 & 68 & & - & - \\
        \cite{Ponza2016}  & 6 & 1,440 &  1,023 &  212 & &  144 &  547 & & - & - \\
        \cite{Ponza2016}  & 7 & 1,440 &  1,893 &  274 & &  196 &  790 & & - & - \\
        \cite{Ponza2016}  & 8 & 1,440 &  4,411 &  344 & &  281 &  1,125 & & - & - \\
        \cite{Ponza2016}  & 9 & 1,440 &  4,161 &  422 & &  295 &  1,439 & & - & - \\
        \cite{Ponza2016}  & 10 & 1,440 &  9,208 &  508 & &  422 &  1,941 & & - & - \\

        \midrule
        \cite{Murray2015} & 10 &  1,200 &  31,758  &  453 & & - & - & &  867 &  2,840\\
        \cite{Murray2015} & 10 &  2,400 &  41,273  &  453 & & - & - &&  867 &  2,840\\
        \bottomrule
    \end{tabular}
    \end{adjustbox}
    \label{formulation_sizes}
\end{table}

The formulation given by \eqref{fo}--\eqref{bin} has a total of $\mathcal{O}(|V|^5)$ variables. 
However, as aforementioned, the number of variables generated is proportional to the sizes of sets $A$ and $D$ (see Section~\ref{sec:MIPModel}), which are generally much smaller in practice than $|V|^2$ and $|V|^3$, respectively.
Table~\ref{formulation_sizes} presents the average number of generated variables  (\#Vars) and constraints  (\#Constrs) for the instances considered.
Note how the formulation's dimensions depend heavily upon the endurance of the drone ($e$).
This is expected since a smaller endurance enables reducing set $D$'s size.
It is also noteworthy how the number of variables actually within the model is not prohibitive for small instances.

\setlength{\tabcolsep}{6pt} 
\renewcommand{\arraystretch}{1.0} 
\begin{table}[b!]
    \footnotesize
    \caption{Formulation results for \cite{Murray2015} instances}
    \label{tab:murray}%
    \begin{adjustbox}{width=1\textwidth}
    \begin{tabular}{llrrrrrrrrrrrrr}
        \toprule
        && \multicolumn{6}{c}{$e = 20$} && \multicolumn{6}{c}{$e = 40$} \\ 
        \cmidrule{3-8} \cmidrule{10-15}
            \multicolumn{1}{c}{Inst.} &
            & \multicolumn{2}{c}{M\&C (2015)} && \multicolumn{3}{c}{This Work} &
            & \multicolumn{2}{c}{M\&C (2015)} && \multicolumn{3}{c}{This Work}\\ 
        \cmidrule{3-4} \cmidrule{6-8} \cmidrule{10-11} \cmidrule{13-15}
            && \multicolumn{1}{c}{Sol.} 
            & \multicolumn{1}{c}{Time} && \multicolumn{1}{c}{LB$^0$} & \multicolumn{1}{c}{Sol.} & \multicolumn{1}{c}{Time} & 
            & \multicolumn{1}{c}{Sol.} & \multicolumn{1}{c}{Time} && \multicolumn{1}{c}{LB$^0$} & \multicolumn{1}{c}{Sol.} & \multicolumn{1}{c}{Time}\\
        \midrule
        \emph{A}-v1   && 56.47 & 1800.15 && 38.78 & {$^\circledast$} 56.47 & 10.73  & & 52.10 & 1800.15 && 31.93 & {$^\circledast$} 50.57 & 1305.59 \\
        \emph{A}-v2   && 53.21 & 1800.15 && 36.05 & {$^\circledast$} 53.21 & 10.07  & & 47.31 & 1800.15 && 27.82 & {$^\circledast$} 47.31 & 671.41 \\
        \emph{A}-v3   && 53.69 & 1800.18 && 37.48 & {$^\circledast$} 53.69 & 10.71  & & 53.69 & 1800.18 && 30.89 & {$^\circledast$} 53.69 & 925.35 \\
        \emph{A}-v4   && 67.46 & 1800.15 && 51.77 & {$^\circledast$} 67.46 & 7.30   & & 66.49 & 1800.15 && 43.66 & {$^\circledast$} 66.49 & 675.25 \\
        \emph{A}-v5   && 50.55 & 1800.22 && 30.67 & {$^\circledast$} 50.55 & 455.30 & & 45.84 & 1800.22 && 30.67 & {$^\circledast$} 44.84 & 764.31 \\
        \emph{A}-v6   && 47.60 & 1800.23 && 27.69 & {$^\circledast$} 47.31 & 330.01 & & 47.60 & 1800.23 && 27.68 & {$^\circledast$} 43.60 & 569.86 \\
        \emph{A}-v7   && 51.89 & 1800.24 && 30.85 & {$^\circledast$} 48.58 & 69.80  & & 46.62 & 1800.24 && 30.84 & {$^\circledast$} 46.62 & 491.20 \\
        \emph{A}-v8   && 64.69 & 1800.23 && 43.60 & {$^\circledast$} 61.38 & 56.82  & & 59.78 & 1800.23 && 43.58 & {$^\circledast$} 59.42 & 677.58 \\
        \emph{A}-v9   && 45.98 & 1800.25 && 30.62 & {$^\circledast$} 42.42 & 92.00  & & 42.42 & 1800.25 && 30.62 & {$^\circledast$} 42.42 & 211.23 \\
        \emph{A}-v10  && 43.09 & 1800.28 && 27.60 & {$^\circledast$} 41.73 & 100.72 & & 41.73 & 1800.28 && 27.60 & {$^\circledast$} 41.73 & 157.21 \\
        \emph{A}-v11  && 48.21 & 1800.25 && 30.81 & {$^\circledast$} 42.90 & 19.26  & & 42.90 & 1800.25 && 30.81 & {$^\circledast$} 42.90 & 142.32 \\
        \emph{A}-v12  && 61.57 & 1800.27 && 43.54 & {$^\circledast$} 55.70 & 28.37  & & 55.70 & 1800.27 && 43.54 & {$^\circledast$} 55.70 & 102.14 \\
        \midrule
        \emph{B}-v1   && 49.43 & 1800.15 && 28.23 & {$^\circledast$} 49.43 & 26.08  & & 48.72 & 1800.15 && 28.22 & {$^\circledast$} 46.89 & 543.02 \\
        \emph{B}-v2   && 50.71 & 1800.15 && 28.22 & {$^\circledast$} 50.71 & 20.48  & & 46.42 & 1800.15 && 28.20 & {$^\circledast$} 46.42 & 137.92 \\
        \emph{B}-v3   && 56.10 & 1800.17 && 35.00 & {$^\circledast$} 56.10 & 21.50  & & 53.93 & 1800.17 && 34.99 & {$^\circledast$} 53.93 & 575.93 \\
        \emph{B}-v4   && 69.90 & 1800.14 && 49.00 & {$^\circledast$} 69.90 & 19.42  & & 68.40 & 1800.14 && 47.76 & {$^\circledast$} 68.40 & 640.12 \\
        \emph{B}-v5   && 45.36 & 1800.22 && 28.17 & {$^\circledast$} 43.53 & 44.28  & & 46.59 & 1800.22 && 28.17 & {$^\circledast$} 43.53 & 101.52 \\
        \emph{B}-v6   && 44.08 & 1800.22 && 27.93 & {$^\circledast$} 43.95 & 40.88  & & 44.08 & 1800.22 && 27.93 & {$^\circledast$} 43.81 & 69.44 \\
        \emph{B}-v7   && 51.92 & 1800.22 && 34.94 & {$^\circledast$} 49.42 & 43.36  & & 49.20 & 1800.22 && 34.94 & {$^\circledast$} 49.20 & 86.90 \\
        \emph{B}-v8   && 65.62 & 1800.22 && 47.74 & {$^\circledast$} 62.22 & 39.82  & & 62.27 & 1800.22 && 47.74 & {$^\circledast$} 62.00 & 38.63 \\
        \emph{B}-v9   && 44.25 & 1800.27 && 28.15 & {$^\circledast$} 42.53 & 62.33  & & 44.25 & 1800.27 && 28.15 & {$^\circledast$} 42.53 & 36.83 \\
        \emph{B}-v10  && 43.08 & 1800.27 && 27.81 & {$^\circledast$} 43.08 & 60.98  & & 43.08 & 1800.27 && 27.81 & {$^\circledast$} 43.08 & 43.33 \\
        \emph{B}-v11  && 49.20 & 1800.27 && 34.93 & {$^\circledast$} 49.20 & 35.41  & & 49.20 & 1800.27 && 34.93 & {$^\circledast$} 49.20 & 71.60 \\
        \emph{B}-v12  && 62.00 & 1800.27 && 47.73 & {$^\circledast$} 62.00 & 54.01  & & 62.00 & 1800.27 && 47.73 & {$^\circledast$} 62.00 & 46.88 \\
        \midrule
        \emph{C}-v1   && 69.59 & 1800.16 && 54.27 & {$^\circledast$} 69.59 & 4.07   & & 57.25 & 1800.16 && 31.29 & {$^\circledast$} 55.49 & 1062.18 \\
        \emph{C}-v2   && 72.15 & 1800.14 && 58.45 & {$^\circledast$} 72.15 & 4.92   & & 58.05 & 1800.14 && 36.29 & {$^\circledast$} 58.05 & 920.71 \\
        \emph{C}-v3   && 77.34 & 1800.13 && 65.44 & {$^\circledast$} 77.34 & 1.90   & & 69.17 & 1800.13 && 49.20 & {$^\circledast$} 68.43 & 436.34 \\
        \emph{C}-v4   && 90.14 & 1800.16 && 78.59 & {$^\circledast$} 90.14 & 3.26   & & 82.70 & 1800.16 && 62.13 & {$^\circledast$} 82.70 & 384.01 \\
        \emph{C}-v5   && 63.25 & 1800.22 && 33.55 & {$^\circledast$} 53.05 & 32.20  & & 53.45 & 1800.22 && 30.92 & {$^\circledast$} 51.93 & 1801.24 \\
        \emph{C}-v6   && 64.70 & 1800.24 && 36.81 & {$^\circledast$} 55.21 & 61.71  & & 52.33 & 1800.24 && 36.29 & {$^\circledast$} 52.33 & 488.63 \\
        \emph{C}-v7   && 67.77 & 1800.21 && 51.37 & {$^\circledast$} 64.41 & 34.90  & & 60.74 & 1800.21 && 49.09 & {$^\circledast$} 60.74 & 86.58 \\
        \emph{C}-v8   && 83.70 & 1800.20 && 64.35 & {$^\circledast$} 77.21 & 32.95  & & 74.69 & 1800.20 && 61.89 & {$^\circledast$} 72.97 & 61.09 \\
        \emph{C}-v9   && 59.32 & 1800.23 && 30.92 & {$^\circledast$} 45.93 & 170.79 & & 47.25 & 1800.23 && 30.92 & {$^\circledast$} 45.93 & 261.44 \\
        \emph{C}-v10  && 61.24 & 1800.23 && 36.29 & {$^\circledast$} 46.93 & 32.20  & & 48.87 & 1800.23 && 36.29 & {$^\circledast$} 46.93 & 48.91 \\
        \emph{C}-v11  && 67.43 & 1800.23 && 49.09 & {$^\circledast$} 56.40 & 19.65  & & 56.40 & 1800.23 && 49.09 & {$^\circledast$} 56.40 & 23.48 \\
        \emph{C}-v12  && 83.70 & 1800.22 && 61.89 & {$^\circledast$} 69.20 & 9.25   & & 69.20 & 1800.22 && 61.89 & {$^\circledast$} 69.20 & 16.04 \\
        \bottomrule
    \end{tabular}
    \end{adjustbox}
\end{table}

Table~\ref{tab:murray} presents the results obtained by the altered formulation considering the instances from \cite{Murray2015} with $e=20$ and $e=40$. For compactness, we refer to \cite{Murray2015} as M\&C (2015), to instance \textit{20140810T123437} as \emph{A}, \textit{20140810T123440} as \emph{B} and \textit{20140810T123443} as \emph{C}. 
Column LB$^\text{0}$ presents the value of the linear relaxation, column $Sol.$ presents the solution value and column Time reports the total execution runtime in seconds. 
Note that a $\circledast$ is included next to column $Sol.$ whenever the solution is proven optimal by the solver using the indicated formulation.
Note also that a runtime limit of 1800 seconds was imposed and that we omitted column LB$^\text{0}$ for the formulation proposed by \cite{Murray2015}, since it obtained value zero for all instances.

The proposed formulation resulted in proven optimal solutions for all instances considering Table~\ref{tab:murray}, taking 61 seconds on average.
This result is quite remarkable since with the formulation proposed by \cite{Murray2015} the solver was incapable of providing any proven optimal solution within the runtime limit.

\begin{table}[b!]
  \centering
  \caption{Formulation results for \cite{Ponza2016} instances}
  \label{tab:ponza}
    \footnotesize
    \begin{tabular}{crrrrrr}
    \toprule
    Instance & \multicolumn{2}{c}{\cite{Ponza2016}} && \multicolumn{3}{c}{This Work} \\
    \cmidrule{2-3} \cmidrule{5-7}
          & \multicolumn{1}{c}{Sol.} & \multicolumn{1}{c}{Time} && \multicolumn{1}{c}{LB$^0$} & \multicolumn{1}{c}{Sol.} & \multicolumn{1}{c}{Time} \\
          \midrule
    Instance\_005.1 & 4456.83 & 0.13  && 3851.22 & {\scriptsize$^\circledast$}4456.83 & 0.38 \\
    Instance\_005.2 & 3507.07 & 0.12  && 1984.71 & {\scriptsize$^\circledast$}3507.07 & 0.07 \\
    Instance\_005.3 & 3275.69 & 0.14  && 2979.03 & {\scriptsize$^\circledast$}3275.69 & 0.12 \\
    Instance\_005.4 & 5312.47 & 0.09  && 3423.66 & {\scriptsize$^\circledast$}5312.47 & 0.07 \\
    Instance\_005.5 & 5510.17 & 0.10  && 5021.23 & {\scriptsize$^\circledast$}5510.17 & 0.05 \\
    \midrule
    Instance\_006.1 & 7080.94 & 0.25  && 6064.16 & {\scriptsize$^\circledast$}7080.94 & 0.08 \\
    Instance\_006.2 & 6147.96 & 0.32  && 5713.98 & {\scriptsize$^\circledast$}6147.96 & 0.23 \\
    Instance\_006.3 & 6835.16 & 0.23  && 5878.56 & {\scriptsize$^\circledast$}6835.16 & 0.08 \\
    Instance\_006.4 & 4402.08 & 0.32  && 3424.12 & {\scriptsize$^\circledast$}4402.08 & 0.41 \\
    Instance\_006.5 & 5392.08 & 0.38  && 4031.53 & {\scriptsize$^\circledast$}5392.08 & 0.34 \\
    \midrule
    Instance\_007.1 & 5533.85 & 3.31  && 3606.98 & {\scriptsize$^\circledast$}5533.85 & 0.48 \\
    Instance\_007.2 & 5342.68 & 1.79  && 3258.57 & {\scriptsize$^\circledast$}5342.68 & 0.96 \\
    Instance\_007.3 & 7725.89 & 1.07  && 6293.13 & {\scriptsize$^\circledast$}7725.89 & 0.21 \\
    Instance\_007.4 & 7610.38 & 1.39  && 6284.05 & {\scriptsize$^\circledast$}7610.38 & 0.16 \\
    Instance\_007.5 & 7010.99 & 2.10  && 6211.52 & {\scriptsize$^\circledast$}7010.99 & 0.27 \\
    \midrule
    Instance\_008.1 & 6709.02 & 5.90  && 4764.75 & {\scriptsize$^\circledast$}6709.02 & 1.26 \\
    Instance\_008.2 & 6587.18 & 10.08 && 4916.63 & {\scriptsize$^\circledast$}6587.18 & 2.06 \\
    Instance\_008.3 & 5780.12 & 14.68 && 4133.14 & {\scriptsize$^\circledast$}5780.12 & 3.00 \\
    Instance\_008.4 & 6505.12 & 8.91  && 3694.07 & {\scriptsize$^\circledast$}6505.12 & 1.76 \\
    Instance\_008.5 & 5953.51 & 15.72 && 4748.36 & {\scriptsize$^\circledast$}5953.51 & 2.48 \\
    \midrule
    Instance\_009.1 & 7338.77 & 189.38 && 5773.50 & {\scriptsize$^\circledast$}7338.77 & 2.95 \\
    Instance\_009.2 & 6204.63 & 129.12 && 4073.60 & {\scriptsize$^\circledast$}6204.63 & 3.30 \\
    Instance\_009.3 & 7698.14 & 87.45 && 3995.14 & {\scriptsize$^\circledast$}7698.14 & 5.16 \\
    Instance\_009.4 & 6817.72 & 79.71 && 4281.48 & {\scriptsize$^\circledast$}6817.72 & 3.69 \\
    Instance\_009.5 & 7802.67 & 115.02 && 5253.94 & {\scriptsize$^\circledast$}7802.67 & 4.85 \\
    \midrule
    Instance\_010.1 & 5986.71 & 1800.15 && 4502.21 & {\scriptsize$^\circledast$}5986.71 & 50.82 \\
    Instance\_010.2 & 6394.39 & 1800.15 && 5141.80 & {\scriptsize$^\circledast$}6394.39 & 15.81 \\
    Instance\_010.3 & 6310.60 & 1800.15 && 3204.10 & {\scriptsize$^\circledast$}6310.60 & 21.94 \\
    Instance\_010.4 & 8377.92 & 752.87 && 7186.84 & {\scriptsize$^\circledast$}8377.92 & 3.58 \\
    Instance\_010.5 & 8934.41 & 1800.15 && 5662.50 & {\scriptsize$^\circledast$}8934.41 & 10.56 \\
    \bottomrule
    \end{tabular}%
\end{table}%

Table \ref{tab:ponza} presents the results obtained by Formulation \eqref{fo}--\eqref{bin} without alterations considering \cite{Ponza2016}'s smaller instances containing from 5 to 10 customers. 
No formulation was capable of solving larger instances with 50, 100, 150 and 200 customers.  
For these instances, the formulations presented by \cite{Ponza2016} did not obtain any feasible solution and the formulation we propose could not be executed due to memory limitations. 
It is thus by no means a coincidence that these large instances have only been addressed with heuristic approaches so far. Therefore, we developed the HTGVNS to tackle the large instances found in the literature.

The formulations proposed by \cite{Ponza2016} also obtained linear relaxation lower bounds of value zero for all instances, and so we omitted column LB$^0$. 
It is clear that the formulations previously proposed in the literature are outperformed by the one we propose, which was capable of producing significantly better lower bounds and by consequence proven optimal solutions for all small instances, which are represented by a {$\circledast$} in the table.

\subsection{HTGVNS Results}\label{sec:heuristic}

In this section, solutions provided by HTGVNS for different benchmarks and the parameters of the heuristic are detailed.

The HTGVNS tabu list size $|TL|$ and perturbation level $\rho_{max}$ were manually tuned.
After a significant number of experiments considering different values for the parameters, $\rho_{max}$ was defined as $\left\lceil n\times\frac{1}{10}\right\rceil$, with $n$ being the number of customers.
Two sets were selected based on their performance to be the tabu list size: $|TL| = 2$ for instances with up to 20 customers and $|TL| = 7$ for instances with more than 20 customers.
The solutions are kept in the tabu list until $|TL|$ moves involving the drone are performed.
The problem-specific input parameters, such as drone endurance, service time and vehicles speed, are defined individually for each instance set (see Sections \ref{sec:ponza} -- \ref{sec:tsplib}).

From all the neighborhood structures implemented, the most effective ones for all set of instances were \emph{Shift(1,0)} and \emph{Swap(1,1)} \emph{intra-route}, which improved the solution by 78\% and 43\% of the times they were considered, respectively. The reinsertion moves, \emph{Reinsertion} and \emph{Or-opt2}, presented the worst performance when analyzed individually.

The tables presented hereafter employ the following notation. Column \textit{Inst.} indicates the instance name, and $BKS$ describes the best-known solution reported in the literature. Columns $Sol.$, $Gap$ and $Time$ indicate, the best solution value, the gap between the best solution found by HTGVNS and BKS and the computational time in seconds, respectively. Finally, columns $\overline{Sol.}$ and $\overline{Gap}$ present the average solution cost of ten runs and the gap between the average solution cost of HTGVNS and the BKS.

HTGVNS uses the optimal TSP solution to create the initial FSTSP solution. The TSP solution is obtained using Concorde solver \citep{Concorde} version 3.12.19 configured with CPLEX 12.6.3. Concorde is capable of finding the optimal solution without considerably increasing HTGVNS computational time. The computational experiments show that Concorde TSP Solver uses less than 7\% of HTGVNS total time.

\subsubsection{Results for Ponza's benchmark set}
\label{sec:ponza}

\cite{Ponza2016} generated 50 instances that can be separated into two sets. A small set contains between 5 to 10 customers, which results can be checked in \label{sec:MurrayModel}, and a larger one including instances with 50, 100, 200 customers. Five instances were generated for each number of customers.

\cite{Ponza2016} found the optimal solution to the smaller instance set by running the Mixed Integer Linear Programming (MILP) presented in his work. We achieved the optimal value in all instances by running HTGVNS ten times for each instance size.
Concorde was not able to find a solution to some instances; therefore, we used the Cheapest Insertion (CI) heuristic to obtain the initial solution for every test problem.
For the larger instances, the initial solution was obtained with Concorde solver. The results are presented in Table \ref{tab:largersetPonza}. The first column indicates the number of customers followed by a complement to keep track of the number of instances per customer. The HTGVNS algorithm was able to improve the solution quality up to 15\% when compared to \cite{Ponza2016} and 6\% compared to \cite{FreitasPenna2020}.
\cite{Ponza2016} implemented a Simulated Annealing (SA) which the average of ten runs are presented in column $Sol$ for \cite{Ponza2016}. Moreover, the computational time concerning \cite{Ponza2016} is the average of all executions for each instance. Therefore, both heuristics has a similar run time.

\setlength{\tabcolsep}{6pt} 
\renewcommand{\arraystretch}{1.0} 
\begin{table}
    \centering
    \small
    \caption{HTGVNS results for \cite{Ponza2016} instances.}
    \label{tab:largersetPonza}%
    \begin{adjustbox}{width=1\textwidth}
    \begin{tabular}{lrrrrrrrrrrrrrrr}
        \toprule
        & & & \multicolumn{2}{c}{SA} && \multicolumn{3}{c}{HGVNS} && \multicolumn{5}{c}{HTGVNS} \\
        & & & \multicolumn{2}{c}{\cite{Ponza2016}} && \multicolumn{3}{c}{F\&P (2020)} && \multicolumn{5}{c}{} \\
          \cmidrule(l){4-5} \cmidrule(l){7-9} \cmidrule(l){11-15}
        & \multicolumn{1}{c}{BKS} && \multicolumn{1}{c}{Sol.} & \multicolumn{1}{c}{Time} && \multicolumn{1}{c}{Sol.} & \multicolumn{1}{c}{$\overline{Sol.}$} & \multicolumn{1}{c}{Time} && \multicolumn{1}{c}{Sol.} & \multicolumn{1}{c}{Gap} & \multicolumn{1}{c}{$\overline{Sol.}$} & \multicolumn{1}{c}{$\overline{Gap}$} & \multicolumn{1}{c}{Time} \\
        \midrule
        050.1 & 11506.50 && 12518.93 & 213.87 && 11506.50 & 11857.02 & 4.39  && 11506.50 & 0.00  & 11979.39 & 4.11  & 6.98 \\
        050.2 & 10964.30 && 12475.14 & 208.36 && 10964.30 & 11049.21 & 4.41  && 10964.30 & 0.00  & 10984.98 & 0.19  & 5.38 \\
        050.3 & 11336.40 && 12664.65 & 191.04 && 11336.40 & 11336.40 & 4.03  && 11094.83 & -2.13 & 11132.43 & -1.80 & 5.14 \\
        050.4 & 10856.40 && 12908.18 & 184.85 && 10856.40 & 11929.50 & 5.02  && 10525.92 & -3.04 & 10692.38 & -1.51 & 5.98 \\
        050.5 & 10486.30 && 12164.83 & 189.86 && 10486.30 & 11034.30 & 4.52  && 10399.02 & -0.83 & 10401.25 & -0.81 & 4.35 \\
        100.1 & 15618.00 && 17974.85 & 267.42 && 15618.00 & 15623.84 & 11.94 && 15618.00 & 0.00  & 15832.94 & 1.38  & 15.49 \\
        100.2 & 14899.20 && 17342.18 & 272.35 && 14899.20 & 15127.50 & 9.43  && 14309.33 & -3.96 & 14319.02 & -3.89 & 14.76 \\
        100.3 & 14524.50 && 17181.88 & 265.45 && 14524.50 & 16074.42 & 10.93 && 14283.50 & -1.66 & 14301.39 & -1.54 & 14.30 \\
        100.4 & 15947.30 && 18538.03 & 266.75 && 15947.30 & 15947.30 & 10.34 && 15598.33 & -2.19 & 15604.28 & -2.15 & 13.67 \\
        150.5 & 14948.50 && 17407.43 & 312.77 && 14948.50 & 15479.22 & 9.94  && 14948.50 & 0.00  & 15048.43 & 0.67  & 17.29 \\
        150.1 & 19828.10 && 22823.38 & 365.04 && 19828.10 & 20069.32 & 20.84 && 19828.10 & 0.00  & 20042.43 & 1.08  & 18.84 \\
        150.2 & 20949.30 && 22549.55 & 383.72 && 20949.30 & 21390.32 & 25.05 && 20949.30 & 0.00  & 21132.30 & 0.87  & 27.44 \\
        150.3 & 22633.30 && 23114.14 & 379.99 && 22633.30 & 23108.49 & 29.97 && 22309.39 & -1.43 & 22619.39 & -0.06 & 21.39 \\
        150.4 & 20400.70 && 22651.00 & 382.67 && 20400.70 & 23390.90 & 25.39 && 20198.38 & -0.99 & 20248.43 & -0.75 & 27.20 \\
        150.5 & 22435.52 && 22807.41 & 384.69 && 22435.52 & 23032.05 & 31.84 && 22435.52 & 0.00  & 22798.66 & 1.62  & 26.78 \\
        200.1 & 25648.33 && 26991.21 & 456.74 && 25648.33 & 25983.49 & 30.44 && 25648.33 & 0.00  & 25700.32 & 0.20  & 34.83 \\
        200.2 & 27632.40 && 27848.14 & 452.88 && 27632.40 & 27985.35 & 71.39 && 25765.21 & -6.76 & 25803.29 & -6.62 & 51.39 \\
        200.3 & 26498.33 && 27143.78 & 510.11 && 26498.33 & 26837.33 & 70.38 && 25093.38 & -5.30 & 25193.48 & -4.92 & 57.98 \\
        200.4 & 28247.92 && 28503.18 & 517.44 && 28247.92 & 28463.95 & 87.98 && 26993.38 & -4.44 & 27003.98 & -4.40 & 73.83 \\
        200.5 & 24987.56 && 27875.87 & 515.30 && 24987.56 & 26357.49 & 69.38 && 24987.56 & 0.00  & 25204.44 & 0.87  & 53.08 \\
        \midrule
        \multicolumn{3}{l}{Avg. Gap and Time} &      &336.07   &&   && 26.88 & & & -1.64 & & -0.87 & 24.81 \\
        \bottomrule
    \end{tabular}%
\end{adjustbox}
\end{table}%

\subsubsection{Results for Agatz benchmark set}  \label{sec:agatz}
A variant of the FSTSP is the Traveling Salesman Problem with Drones (TSP-D) introduced by \cite{Agatz2018}.
While, the FSTSP defines endurance ($e$) and service time ($s^L$ and $s^R$), the TSP-D negligence these variable  ($e = \infty$, $s^L = s^R = 0$), therefore, the drone can visit all customers. Furthermore, a drone trip can start and end at the same node (\textit{launch} $=$ \textit{return}). The truck waits for the drone at the \textit{launch} node or visits other customers then returns to the \textit{launch} node.
According to the authors, this strategy is beneficial, considering the truck can visit a node for either serve a customer, or supply the drone with another parcel.
The authors proposed an enormous number of instances to the problem. In \cite{Bouman2015} it is explained how they are generated. They are divided into three sets, \textit{uniform}, \textit{single-center} and \textit{double-center} distribution.
The parameter $\alpha$ defines the ratio of drone speed to truck speed in an instance. Variable $\alpha = 1$ determines that the vehicles travel with the same speed, when $\alpha=2$ the drone travels two times faster than the truck. Lastly, for $\alpha=3$, the drone speed is three times as fast as the truck.

\begin{table}[H]
  \footnotesize
  \centering
  \caption{HTGVNS results for \cite{Agatz2018} instances with uniform distribution}
  \begin{adjustbox}{width=1\textwidth}
    \begin{tabular}{ccrrrrrrrrrr}
    \toprule
          &       &       & \multicolumn{3}{c}{LS} & \multicolumn{3}{c}{HGVNS} & \multicolumn{3}{c}{HTGVNS} \\
          & & & \multicolumn{3}{l}{\citep{Agatz2018}} &   \multicolumn{3}{l}{\citep{FreitasPenna2020}} &   \multicolumn{3}{c}{}   \\
          \cmidrule(l){4-6} \cmidrule(l){7-9} \cmidrule(l){10-12}

          & \multicolumn{1}{c}{n}     & \multicolumn{1}{c}{BKS} & \multicolumn{1}{c}{$\overline{Sol.}$} & \multicolumn{1}{c}{Gap.} & \multicolumn{1}{c}{Time} & \multicolumn{1}{c}{$\overline{Sol.}$} & \multicolumn{1}{c}{Gap.} & \multicolumn{1}{c}{Time} & \multicolumn{1}{c}{$\overline{Sol.}$} & \multicolumn{1}{c}{Gap.} & \multicolumn{1}{c}{Time} \\
          \midrule
    \multirow{8}[0]{*}{\begin{sideways}$\alpha = 1$\end{sideways}} & 10    & 286.82 & \textbf{286.82} & \textbf{0.00}  & 0.00  & 289.82 & 1.05  & 0.14  & 289.82 & 1.05  & 0.15 \\
          & 20    & 365.38 & \textbf{365.38} & \textbf{0.00}  & 0.00  & 368.54 & 0.86  & 0.11  & \textbf{365.38} & \textbf{0.00}   & 0.11 \\
          & 50    & 550.38 & \textbf{550.38} & \textbf{0.00}  & 0.90  & 559.20 & 1.60  & 3.71  & 551.49 & 0.20  & 3.53 \\
          & 75    & 624.32 & 645.55 & 3.40  & 3.20  & \textbf{624.32} & \textbf{0.00}  & 16.30 & \underline{\textbf{624.26}} & \textbf{-0.01} & 16.30 \\
          & 100   & 698.42 & 729.44 & 4.44  & 10.00 & \textbf{698.42} & \textbf{0.00}  & 53.50 & \underline{\textbf{696.15}} & \textbf{-0.33} & 53.75 \\
          & 175   & 905.32 & 940.35 & 3.87  & 98.00 & \textbf{905.32} & \textbf{0.00}  & 55.91 & \underline{\textbf{903.89}} & \textbf{-0.16} & 57.14 \\
          & 250   & 1113.96 & \textbf{1113.96} & \textbf{0.00}  & 410.10 & 1135.32 & 1.92  & 185.19 & \underline{\textbf{1078.00}} & \textbf{-3.23} & 185.19 \\
          & Average &       &       & 1.67  & 74.60 &       & 0.78  & 44.98 &       & -0.35 & 45.17 \\
          \midrule
    \multirow{8}[0]{*}{\begin{sideways}$\alpha = 2$\end{sideways}} & 10    & 231.29 & \textbf{231.29} & \textbf{0.00}  & 0.00  & 233.20 & 0.83  & 0.13  & \underline{\textbf{228.00}} & \textbf{-1.42} & 0.13 \\
          & 20    & 293.59 & \textbf{293.59} & \textbf{0.00}  & 0.00  & \textbf{293.60} & \textbf{0.00}  & 0.85  & \textbf{293.60} & \textbf{0.00}   & 0.85 \\
          & 50    & 420.80 & 428.63 & 1.86  & 1.20  & \textbf{420.80} & \textbf{0.00}  & 2.30  & \textbf{420.80} & \textbf{0.00}   & 2.32 \\
          & 75    & 490.43 & 495.90 & 1.12  & 6.20  & \textbf{490.43} & \textbf{0.00}  & 10.93 & \underline{\textbf{459.71}} & \textbf{-6.26} & 10.95 \\
          & 100   & 553.43 & 572.53 & 3.45  & 18.40 & \textbf{553.43} & \textbf{0.00}  & 37.77 & \textbf{553.43} & \textbf{0.00}   & 37.85 \\
          & 175   & 704.53 & 722.83 & 2.60  & 177.20 & \textbf{704.53} & \textbf{0.00}  & 39.28 & \textbf{704.53} & \textbf{0.00}   & 40.65 \\
          & 250   & 824.42 & 854.34 & 3.63  & 746.90 & \textbf{824.42} & \textbf{0.00}  & 191.48 & \textbf{824.42} & \textbf{0.00}   & 191.48 \\
          & Average &       &       & 1.81  & 135.70 &       & 0.12  & 40.39 &       & -1.10 & 40.60 \\
         \midrule
    \multirow{8}[0]{*}{\begin{sideways}$\alpha = 3$\end{sideways}} & 10    & 210.42 & \textbf{210.42} & \textbf{0.00}  & 0.00  & 215.88 & 2.59  & 0.13  & 211.49 & 0.51  & 0.14 \\
          & 20    & 266.12 & \textbf{266.12} & \textbf{0.00}  & 0.00  & 274.20 & 3.04  & 1.15  & 267.03 & 0.34  & 1.15 \\
          & 50    & 389.80 & 391.96 & 0.55  & 1.60  & \textbf{389.80} & \textbf{0.00}  & 2.19  & \underline{\textbf{377.28}} & \textbf{-3.21} & 2.18 \\
          & 75    & 447.64 & 453.30 & 1.26  & 8.00  & \textbf{447.64} & \textbf{0.00}  & 10.95 & \underline{\textbf{443.30}} & \textbf{-0.97} & 11.05 \\
          & 100   & 510.20 & 530.53 & 3.98  & 25.67 & \textbf{510.20} & \textbf{0.00}  & 37.26 & \textbf{510.20} & \textbf{0.00}  & 37.13 \\
          & 175   & 655.20 & 665.72 & 1.61  & 259.30 & \textbf{655.20} & \textbf{0.00}  & 41.49 & \textbf{655.20} & \textbf{0.00}   & 41.26 \\
          & 250   & 758.32 & 785.86 & 3.63  & 1080.40 & \textbf{758.32} & \textbf{0.00}  & 189.43 & \textbf{758.32} & \textbf{0.00}   & 189.43 \\
          & Average &       &       & 1.58  & 196.42 &       & 0.80  & 40.37 &       & -0.48 & 40.34 \\
          \bottomrule
    \end{tabular}%
    \end{adjustbox}
  \label{table:uniform}%
\end{table}%

Tables \ref{table:uniform} -- \ref{table:doublecenter} are broken down into scenarios stratified by the value of $\alpha$.
The results presented by these tables were collected by running the available code of the author's repository\footnote{https://github.com/pcbouman-eur/Drones-TSP}. It reflects the average of 10 runs for each instance size.
As parameters, we adopted the same configuration employed in HTGVNS to obtain a comparable result.
The TSP initial solution was obtained by the TSP solver Concorde. The four neighborhood structures available were applied: 2-point move (2p) that swaps two nodes in the truck route, 2-opt move where two edges are removed and replaced with two new edges and the 1-point-move (1p) move where a node in the truck route is relocated to a new position. Finally, the last neighborhood combines all the previous moves.

According to the tables, it is possible to notice that although the HTGVNS could not improve the average solution of $\alpha=3$ for double-center distribution this set achieved the best improvement compared to the others distributions and values of $\alpha$.  Concerning computational time, the different distribution of customers does not affect runtime; however, when the vehicles present the same speed runtime increases. \cite{Agatz2018} exact partitioning algorithm outperforms HGVNS computational time for instances up to 100 customers. The scenario changes when the number of customers increases as HTGVNS presents a runtime much smaller than the LS.

\begin{figure}[H]
  \centering
  \caption{Results for \cite{Agatz2018} instances with uniform distribution; left boxplots represent HGVNS results while right boxplots present HTGVNS results.}
  \begin{subfigure}[b]{.47\linewidth}
    \centering
    \includegraphics[width=\linewidth]{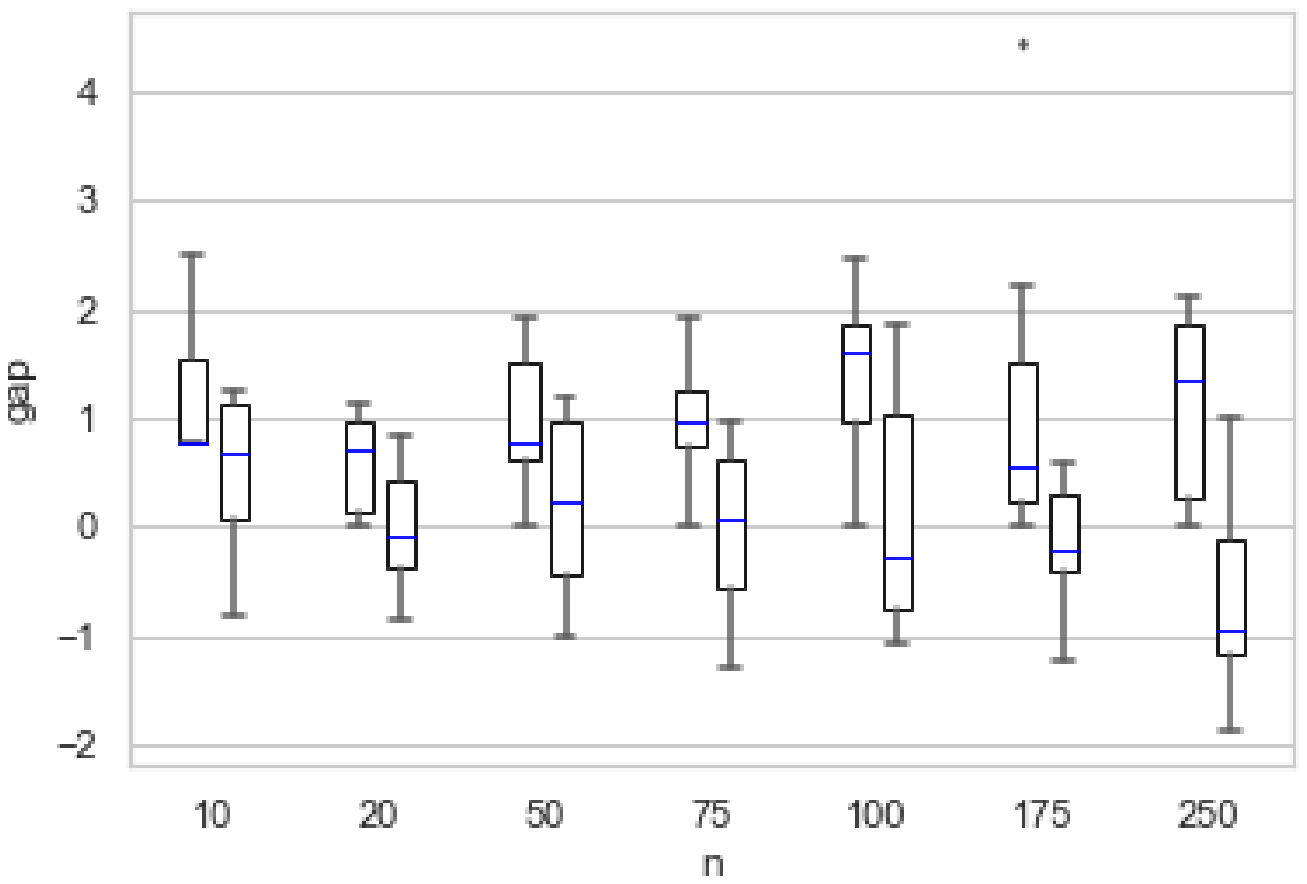}
    \setcounter{subfigure}{0}%
    \caption{$\alpha$ = 1}
  \end{subfigure}
  \begin{subfigure}[b]{.47\linewidth}
    \includegraphics[width=\linewidth]{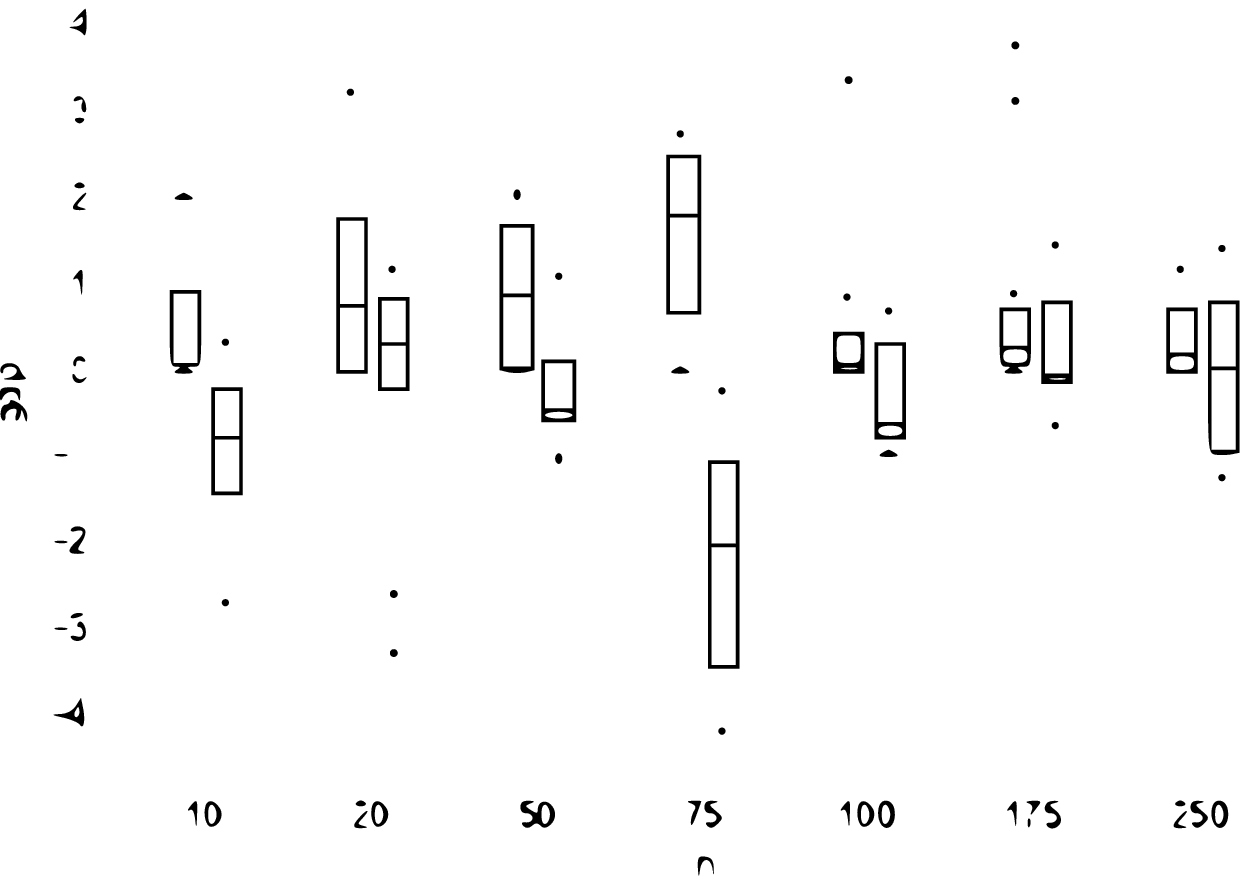}
    \setcounter{subfigure}{1}%
    \caption{$\alpha$ = 2}
  \end{subfigure}
  \begin{subfigure}[b]{.47\linewidth}
    \includegraphics[width=\linewidth]{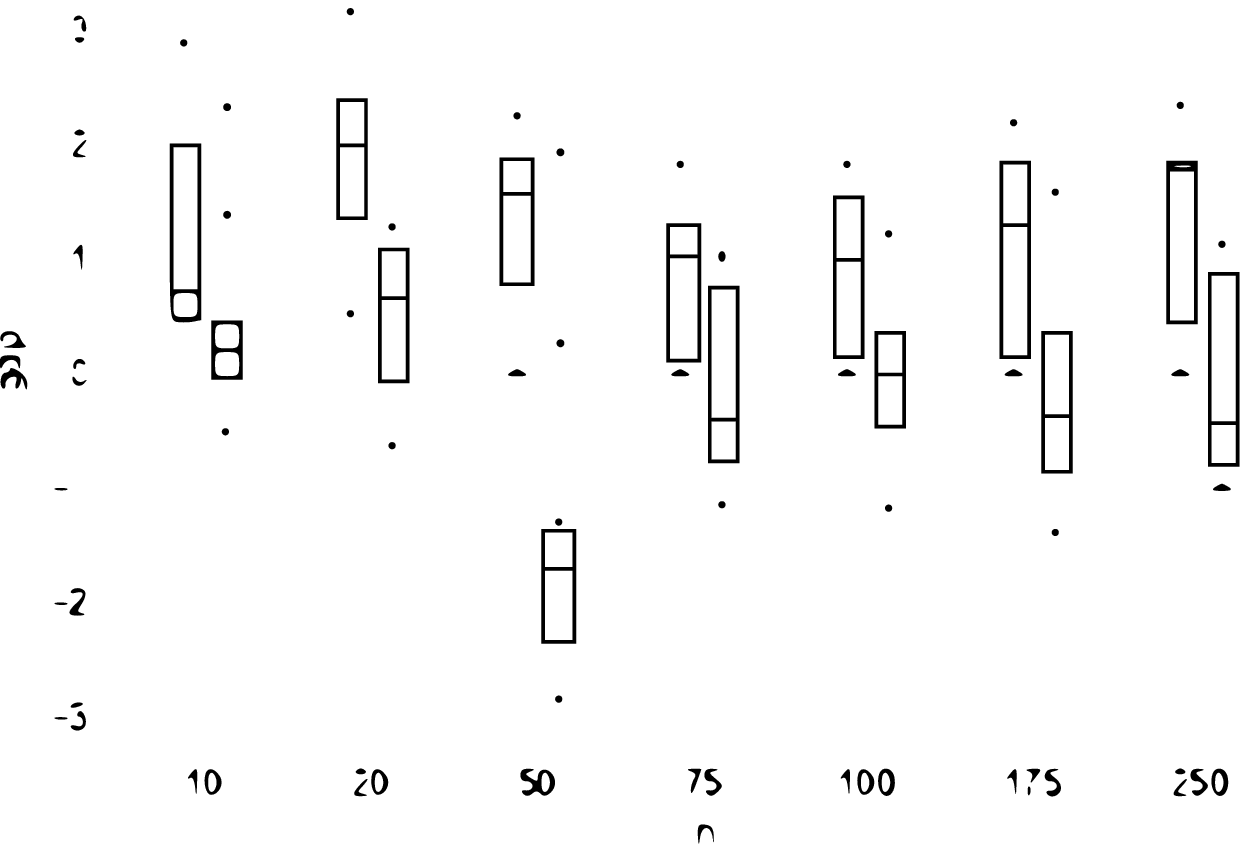}
    \setcounter{subfigure}{1}%
    \caption{$\alpha$ = 3}
  \end{subfigure}
    \label{fig:boxplot_uniform}
\end{figure}

\begin{figure}[H]
  \centering
  \caption{Results for \cite{Agatz2018} instances with single-center distribution; left boxplots represent HGVNS results while right boxplots present HTGVNS results.}
  \begin{subfigure}[b]{.47\linewidth}
    \centering
    \includegraphics[width=\linewidth]{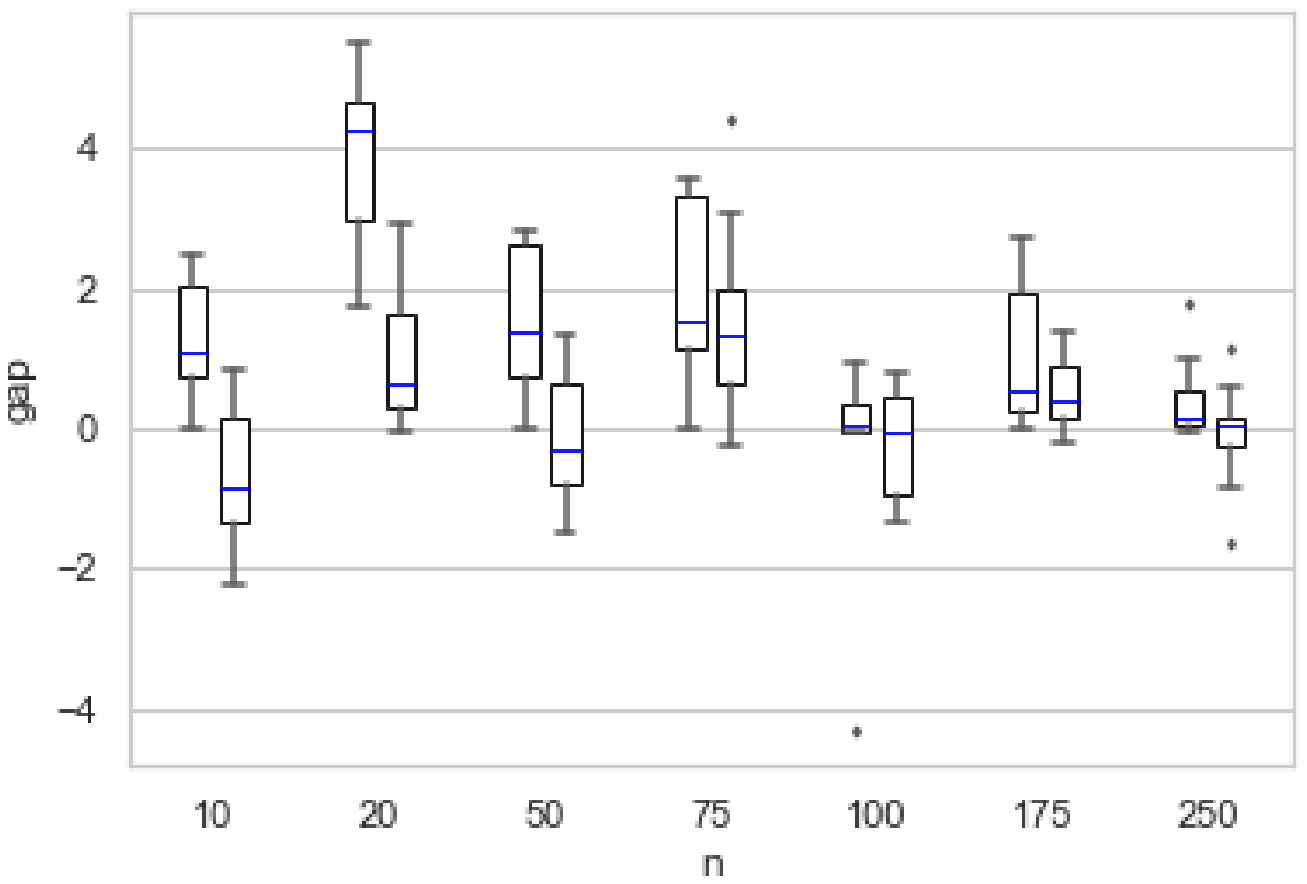}
    \setcounter{subfigure}{0}%
    \caption{$\alpha$ = 1}
  \end{subfigure}
  \begin{subfigure}[b]{.47\linewidth}
    \includegraphics[width=\linewidth]{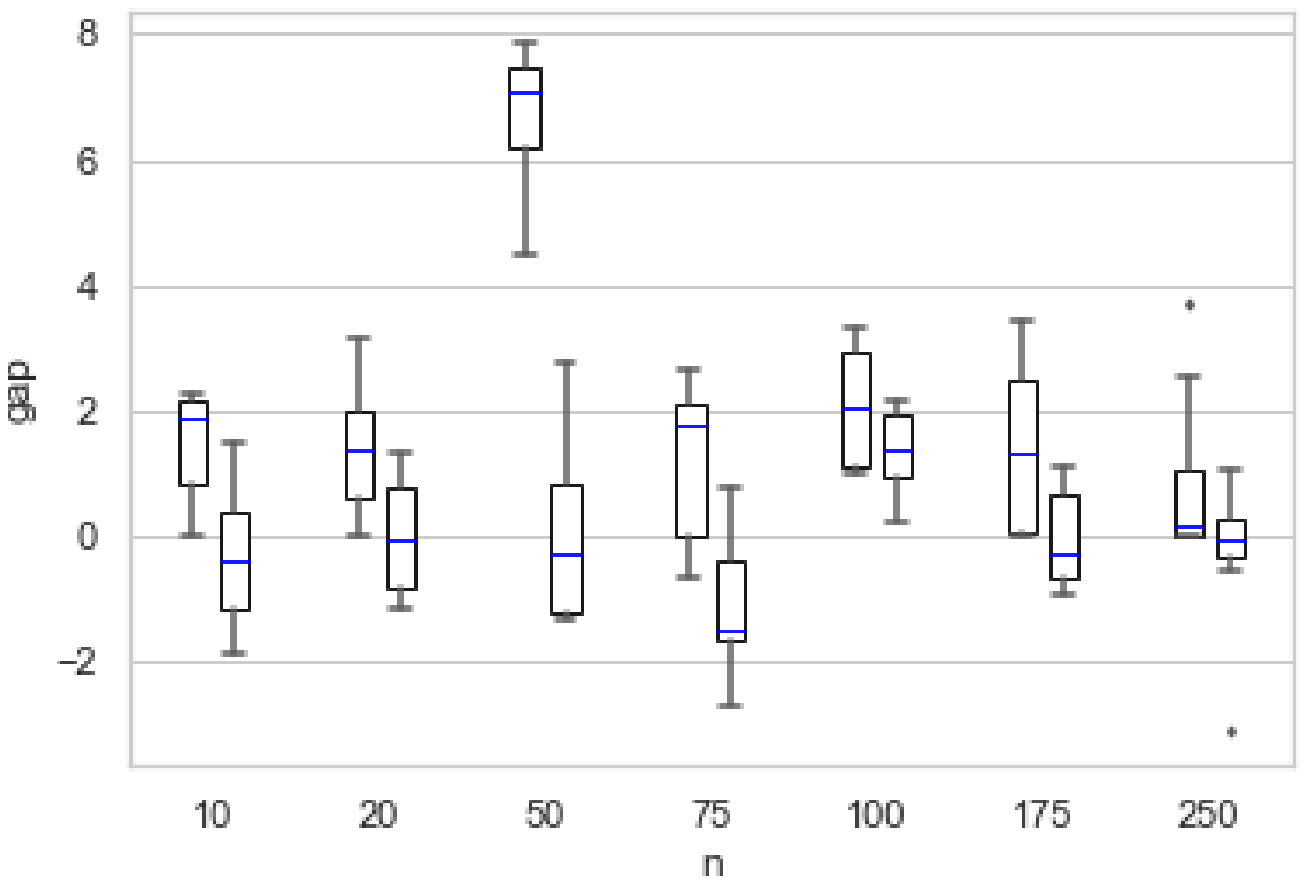}
    \setcounter{subfigure}{1}%
    \caption{$\alpha$ = 2}
  \end{subfigure}
  \begin{subfigure}[b]{.47\linewidth}
    \includegraphics[width=\linewidth]{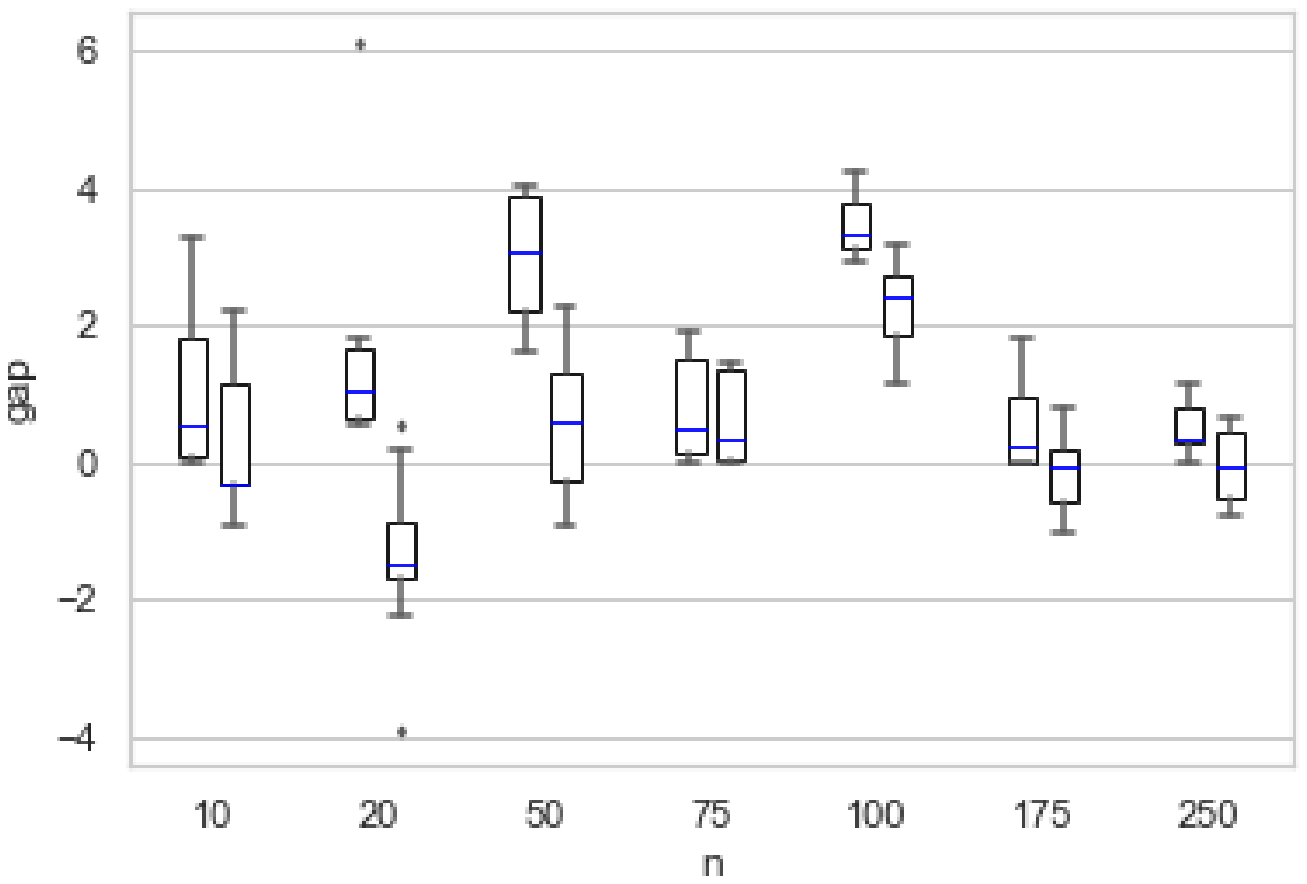}
    \setcounter{subfigure}{1}%
    \caption{$\alpha$ = 3}
  \end{subfigure}
    \label{fig:boxplot_singlecenter}
\end{figure}

\begin{figure}[H]
  \centering
  \caption{Results for \cite{Agatz2018} instances with double-center distribution; left boxplots represent HGVNS results while right boxplots present HTGVNS results.}
  \begin{subfigure}[b]{.47\linewidth}
    \centering
    \includegraphics[width=\linewidth]{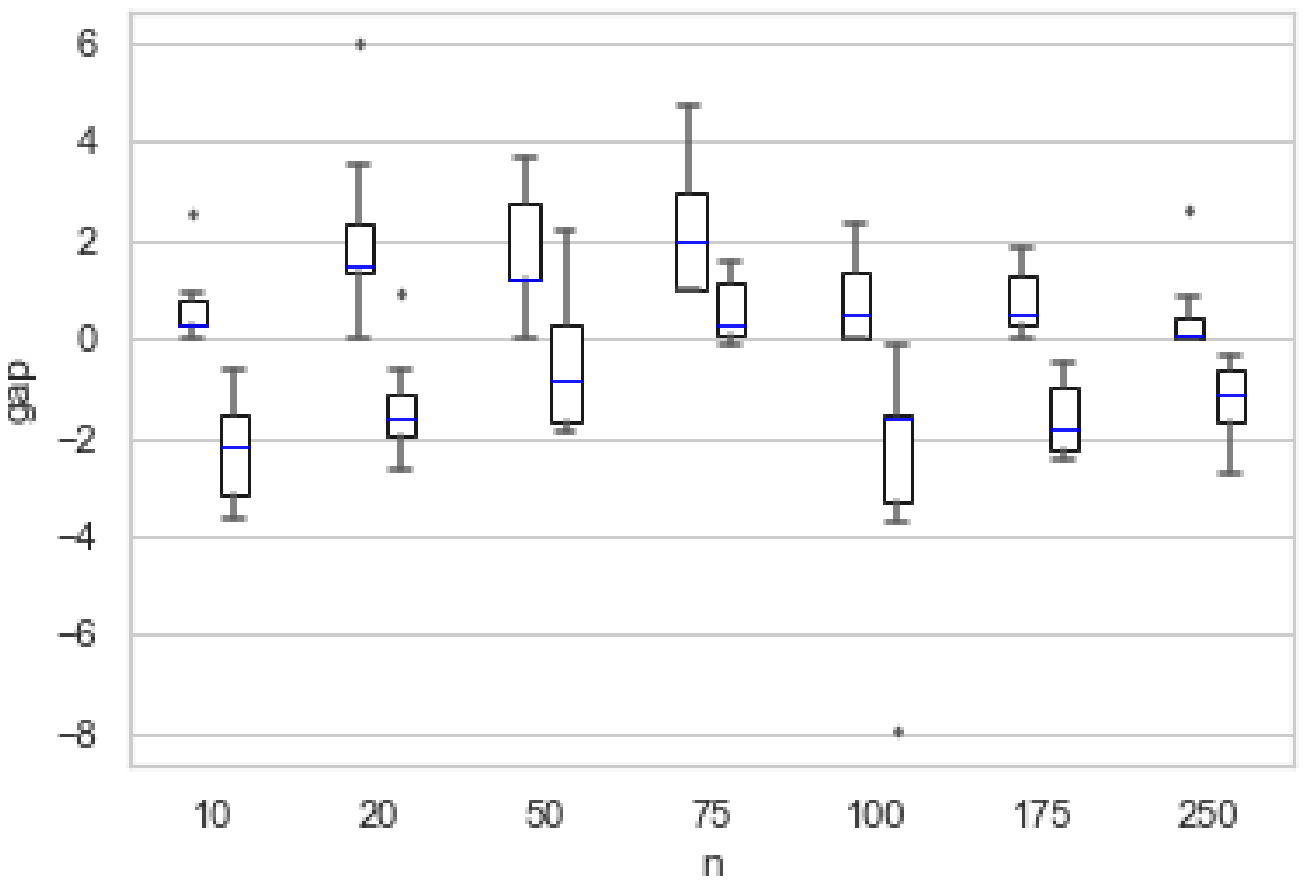}
    \setcounter{subfigure}{0}%
    \caption{$\alpha$ = 1}
  \end{subfigure}
  \begin{subfigure}[b]{.47\linewidth}
    \includegraphics[width=\linewidth]{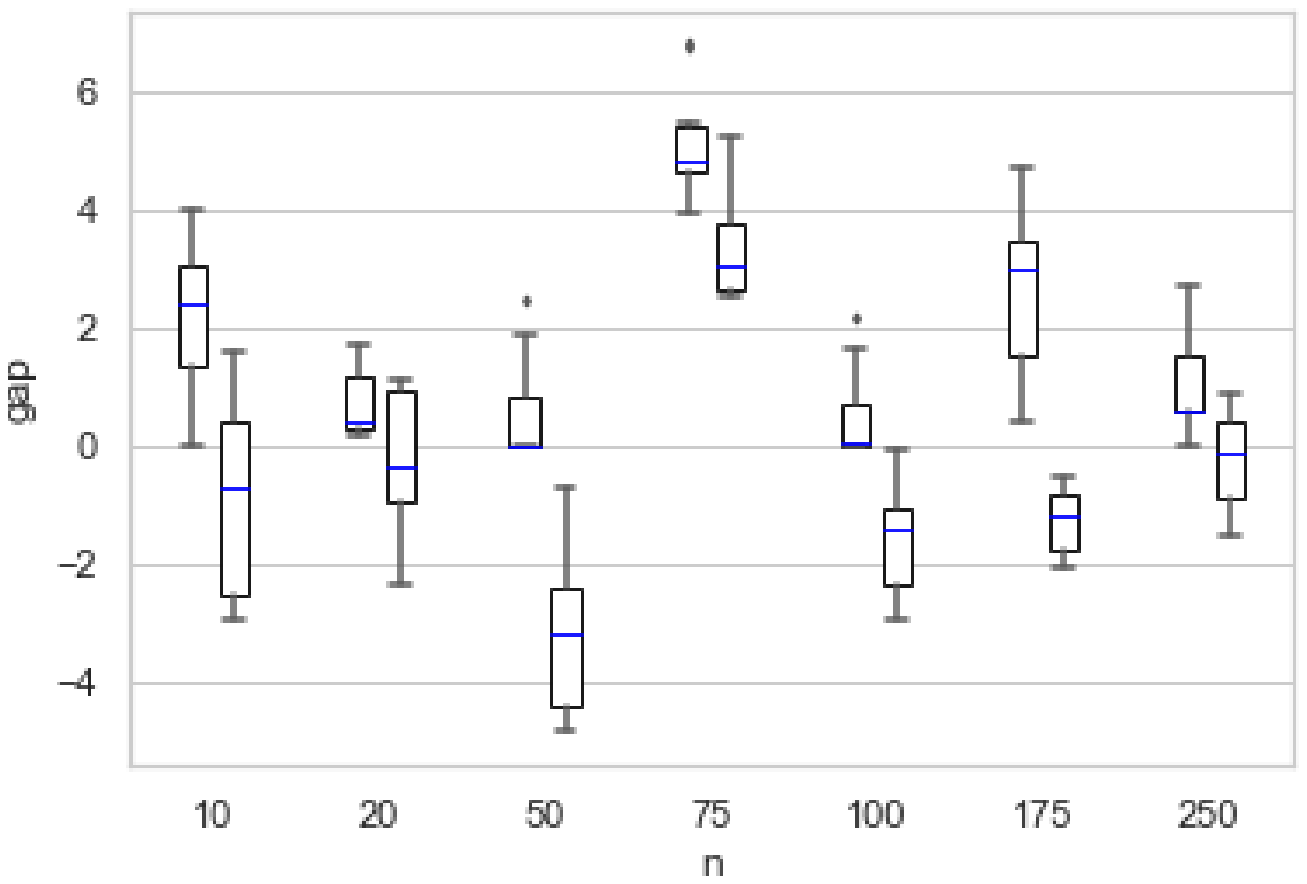}
    \setcounter{subfigure}{1}%
    \caption{$\alpha$ = 2}
  \end{subfigure}
  \begin{subfigure}[b]{.47\linewidth}
    \includegraphics[width=\linewidth]{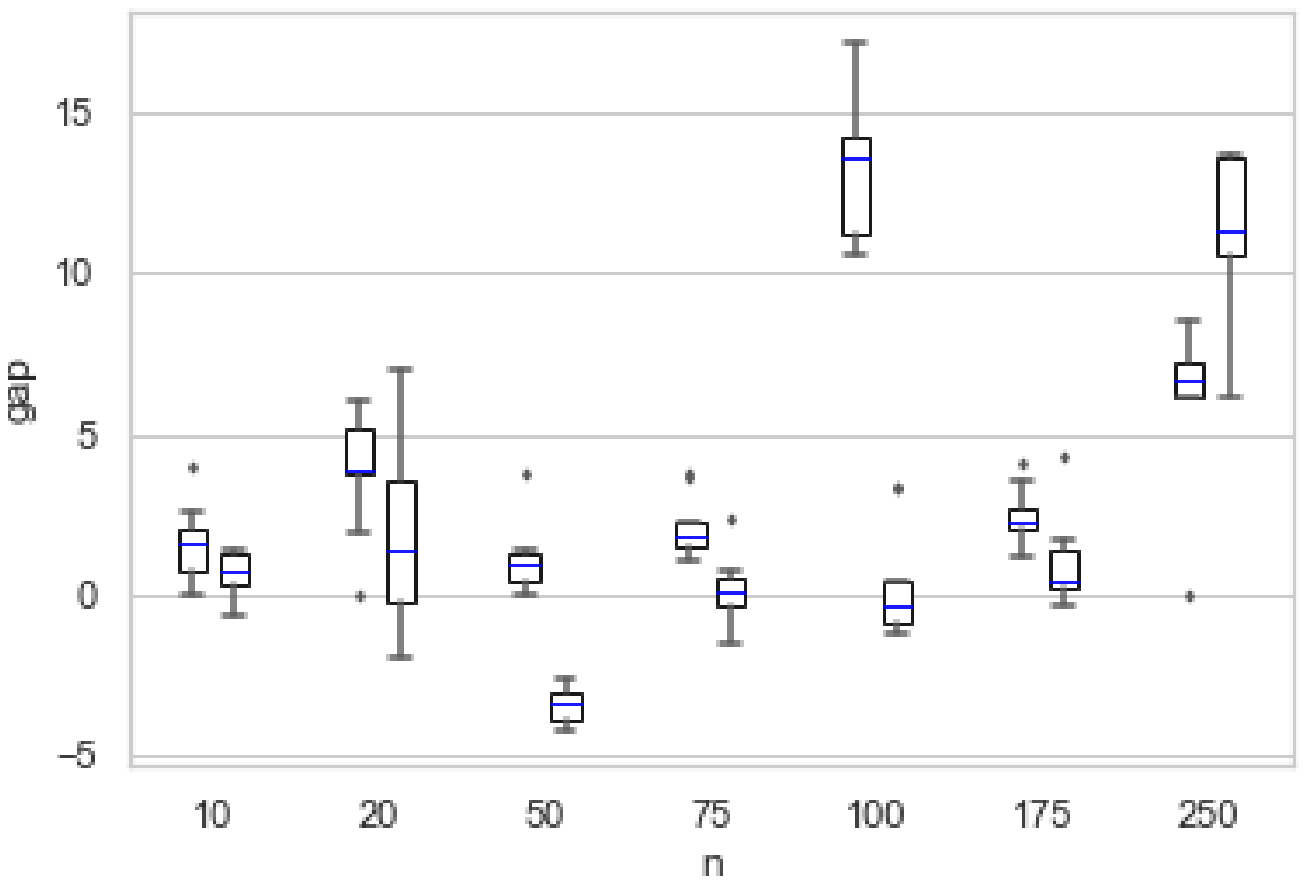}
    \setcounter{subfigure}{1}%
    \caption{$\alpha$ = 3}
  \end{subfigure}
    \label{fig:boxplot_doublecenter}
\end{figure}

Figure \ref{fig:boxplot_uniform}, \ref{fig:boxplot_singlecenter}, and \ref{fig:boxplot_doublecenter} show boxplot graphics within the solution value of 10 runs of HGVNS and HTGVNS. For each distribution (uniform, single-center, and double-center) exist one boxplot representing the instance size (10, 20, 50, 75, 100, 175, 250) of a certain $\alpha$ value.
The thick line corresponds to the median, i.e., half of the values are below this line, divided into two quartiles, and the other half is above this line, dived into two quartiles. The crosses in the graph correspond to the outliers, values that present a considerable distance from the others.
The plots show the tendency of the medium size instance in all distributions have a smaller gap between the minimal and maximum values. It is possible to notice that HTGVNS found better gaps than HGVNS; however, in some instances the gap is still considerable. Therefore, the algorithm still has a lot to be improved.

\begin{table}[H]
  \footnotesize
  \centering
  \caption{Results running HTGVNS in \cite{Agatz2018} with single-center distribution}
  \begin{adjustbox}{width=1\textwidth}
    \begin{tabular}{crrrrrrrrrrr}
            \toprule
          &       &       & \multicolumn{3}{c}{LS} & \multicolumn{3}{c}{HGVNS} & \multicolumn{3}{c}{HTGVNS} \\
          & & & \multicolumn{3}{l}{\citep{Agatz2018}} &   \multicolumn{3}{l}{\citep{FreitasPenna2020}} &   \multicolumn{3}{c}{}   \\
          \cmidrule(l){4-6} \cmidrule(l){7-9} \cmidrule(l){10-12}

          & \multicolumn{1}{c}{n}     & \multicolumn{1}{c}{BKS} & \multicolumn{1}{c}{$\overline{Sol.}$} & \multicolumn{1}{c}{Gap.} & \multicolumn{1}{c}{Time} & \multicolumn{1}{c}{$\overline{Sol.}$} & \multicolumn{1}{c}{Gap.} & \multicolumn{1}{c}{Time} & \multicolumn{1}{c}{$\overline{Sol.}$} & \multicolumn{1}{c}{Gap.} & \multicolumn{1}{c}{Time} \\
          \midrule
    \multirow{8}[0]{*}{\begin{sideways}$\alpha$ = 1\end{sideways}} & \multicolumn{1}{c}{10} & 364.92 & 379.29 & 3.94  & 0.00  & \textbf{364.92} & \textbf{0.00}  & 0.14  & \underline{\textbf{360.85}} & \textbf{-1.11} & 0.14 \\
          & \multicolumn{1}{c}{20} & 529.15 & \textbf{529.15} & \textbf{0.00}  & 0.00  & 553.53 & 4.61  & 1.12  & 537.24 & 1.67 & 1.14 \\
          & \multicolumn{1}{c}{50} & 763.28 & \textbf{763.28} & \textbf{0.00}  & 0.30  & 784.32 & 2.76  & 3.93  & \underline{\textbf{756.20}} & \textbf{-0.93} & 3.72 \\
          & \multicolumn{1}{c}{75} & 978.32 & 1017.04 & 3.96  & 2.80  & \textbf{978.32} & \textbf{0.00}  & 15.32 & \textbf{978.32} & \textbf{0.00}  & 15.50 \\
          & \multicolumn{1}{c}{100} & 1193.95 & 1203.42 & 0.79  & 8.50  & \textbf{1193.95} & \textbf{0.00}  & 55.62 & \textbf{1193.95} & \textbf{0.00}  & 56.64 \\
          & \multicolumn{1}{c}{175} & 1629.32 & 1631.31 & 0.12  & 73.70 & \textbf{1629.32} & \textbf{0.00}  & 53.34 & \textbf{1629.32} & \textbf{0.00}  & 53.55 \\
          & \multicolumn{1}{c}{250} & 1813.54 & 1857.16 & 2.41  & 358.40 & \textbf{1813.54} & \textbf{0.00}  & 249.92 & \textbf{1813.54} & \textbf{0.00}  & 229.92 \\
          & Average &       &       & 1.60  & 63.39 &       & 1.05  & 54.20 &       & -0.68 & 51.51 \\
          \midrule
    \multirow{8}[0]{*}{\begin{sideways}$\alpha$ = 2\end{sideways}} & \multicolumn{1}{c}{10} & 278.22 & \textbf{278.22} & \textbf{0.00}  & 0.00  & 291.36 & 4.72  & 0.14  & 276.95 & -0.46 & 0.14 \\
          & \multicolumn{1}{c}{20} & 364.08 & 384.87 & 5.71  & 0.00  & \textbf{364.08} & \textbf{0.00}  & 1.04  & \underline{\textbf{363.46}} & \textbf{-0.17} & 1.04 \\
          & \multicolumn{1}{c}{50} & 554.58 & \textbf{554.58} & \textbf{0.00}  & 1.00  & 593.54 & 7.03  & 2.23  & \underline{\textbf{553.53}} & \textbf{-0.19} & 2.24 \\
          & \multicolumn{1}{c}{75} & 741.38 & \textbf{741.38} & \textbf{0.00}  & 126.00 & 754.43 & 1.76  & 11.18 & \underline{\textbf{730.43}} & \textbf{-1.48} & 43.36 \\
          & \multicolumn{1}{c}{100} & 891.28 & \textbf{891.28} & \textbf{0.00}  & 5.30  & 900.12 & 0.99  & 38.23 & 908.29 & 1.91  & 11.36 \\
          & \multicolumn{1}{c}{175} & 1183.43 & 1208.94 & 2.16  & 15.70 & \textbf{1183.43} & \textbf{0.00}  & 43.06 & \underline{\textbf{1183.12}} & \textbf{-0.03} & 38.23 \\
          & \multicolumn{1}{c}{250} & 1294.43 & 1396.24 & 7.87  & 136.40 & \textbf{1294.43} & \textbf{0.00}  & 197.12 & \underline{\textbf{1290.38}} & \textbf{-0.31} & 143.06 \\
          & Average &       &       & 2.25  & 40.63 &       & 2.07  & 41.86 &       & \underline{\textbf{-0.10}} & 34.20 \\
          \midrule
    \multirow{8}[0]{*}{\begin{sideways}$\alpha$ = 3\end{sideways}} & \multicolumn{1}{c}{10} & 238.20 & \textbf{238.20} & \textbf{0.00}  & 0.00  & 242.20 & 1.68  & 0.13  & \underline{\textbf{237.43}} & \textbf{-0.32} & 0.13 \\
          & \multicolumn{1}{c}{20} & 315.27 & \textbf{315.27} & \textbf{0.00}  & 0.00  & 319.80 & 1.44  & 0.86  & \underline{\textbf{310.67}} & \textbf{-1.46} & 0.92 \\
          & \multicolumn{1}{c}{50} & 474.82 & \textbf{474.82} & \textbf{0.00}  & 1.00  & 493.43 & 3.92  & 2.19  & 478.32 & 0.74  & 2.26 \\
          & \multicolumn{1}{c}{75} & 638.20 & 639.67 & 0.23  & 7.30  &\textbf{ 638.20} & \textbf{0.00}  & 11.39 & 638.48 & 0.04  & 11.41 \\
          & \multicolumn{1}{c}{100} & 777.25 & \textbf{777.25} & \textbf{0.00}  & 20.80 & 804.43 & 3.50  & 37.74 & 798.54 & 2.74  & 38.40 \\
          & \multicolumn{1}{c}{175} & 998.53 & 1067.54 & 6.91  & 211.60 & \textbf{998.53} & \textbf{0.00}  & 41.20 & \underline{\textbf{990.42}} & \textbf{-0.81} & 41.19 \\
          & \multicolumn{1}{c}{250} & 1224.43 & 1232.56 & 0.66  & 883.80 & \textbf{1224.43} & \textbf{0.00} & 198.72 & \underline{\textbf{1224.13}} & \underline{\textbf{-0.02}} & 198.72 \\
          & Average &       &       & 1.02  & 160.64 &       & 1.41  & 41.75 &       & 0.03  & 41.86 \\
          \bottomrule
    \end{tabular}%
    \end{adjustbox}
  \label{table:singlecenter}%
\end{table}%

\begin{table}[H]
  \footnotesize
  \centering
  \caption{Results running HTGVNS in \cite{Agatz2018} with double-center distribution}
  \begin{adjustbox}{max width=\textwidth}
    \begin{tabular}{rrrrrrrrrrrr}
            \toprule
          &       &       & \multicolumn{3}{c}{LS} & \multicolumn{3}{c}{HGVNS} & \multicolumn{3}{c}{HTGVNS} \\
          & & & \multicolumn{3}{l}{\citep{Agatz2018}} &   \multicolumn{3}{l}{\citep{FreitasPenna2020}} &   \multicolumn{3}{c}{}   \\
          \cmidrule(l){4-6} \cmidrule(l){7-9} \cmidrule(l){10-12}

          & \multicolumn{1}{c}{n}     & \multicolumn{1}{c}{BKS} & \multicolumn{1}{c}{$\overline{Sol.}$} & \multicolumn{1}{c}{Gap.} & \multicolumn{1}{c}{Time} & \multicolumn{1}{c}{$\overline{Sol.}$} & \multicolumn{1}{c}{Gap.} & \multicolumn{1}{c}{Time} & \multicolumn{1}{c}{$\overline{Sol.}$} & \multicolumn{1}{c}{Gap.} & \multicolumn{1}{c}{Time} \\
          \midrule
    \multicolumn{1}{c}{\multirow{8}[0]{*}{\begin{sideways}$\alpha$ = 1\end{sideways}}}
          & \multicolumn{1}{c}{10} & 634.42 & 652.30 & 2.82  & 0.00  & \textbf{634.42} & \textbf{0.00}  & 0.15  & \underline{\textbf{615.80}} & \textbf{-2.93} & 0.09 \\
          & \multicolumn{1}{c}{20} & 754.81 & 776.32 & 2.85  & 0.00  & \textbf{754.81} & \textbf{0.00}  & 1.39  & \underline{\textbf{733.88}} & \textbf{-2.77} & 2.53 \\
          & \multicolumn{1}{c}{50} & 1149.84 & \textbf{1149.84} & \textbf{0.00}  & 0.40  & 1203.09 & 4.63  & 4.34  & 1153.29 & 0.30  & 4.95 \\
          & \multicolumn{1}{c}{75} & 1430.96 & \textbf{1430.96} & \textbf{0.00}  & 3.00  & 1494.57 & 4.45  & 19.39 & 1432.36 & 0.10  & 21.40 \\
          & \multicolumn{1}{c}{100} & 1556.52 & 1605.50 & 3.15  & 9.60  & \textbf{1556.52} & \textbf{0.00}  & 66.39 & \underline{\textbf{1516.36}} & \textbf{-2.58} & 65.30 \\
          & \multicolumn{1}{c}{175} & 2072.36 & 2155.85 & 4.03  & 88.30 & \textbf{2072.36} & \textbf{0.00}  & 57.03 & \underline{\textbf{2030.44}} & \textbf{-2.02} & 54.33 \\
          & \multicolumn{1}{c}{250} & 2523.00 & 2572.80 & 1.97  & 353.60 & \textbf{2523.00} & \textbf{0.00}  & 260.98 & \underline{\textbf{2493.64}} & \textbf{-1.16} & 273.44 \\
          & \multicolumn{1}{c}{Average} &       &       & 2.12  & 64.99 &       & 1.30  & 58.53 &       & -1.58 & 60.29 \\
          \midrule
    \multicolumn{1}{c}{\multirow{8}[0]{*}{\begin{sideways}$\alpha$ = 2\end{sideways}}}
          & \multicolumn{1}{c}{10} & 490.64 & \textbf{490.64} & \textbf{0.00}  & 0.00  & 510.09 & 3.96  & 0.13  & \underline{\textbf{487.24}} & \textbf{-0.69} & 0.33 \\
          & \multicolumn{1}{c}{20} & 603.31 & \textbf{603.31} & \textbf{0.00}  & 0.00  & 605.04 & 0.29  & 0.98  & \underline{\textbf{602.73}} & \textbf{-0.10} & 1.42 \\
          & \multicolumn{1}{c}{50} & 850.31 & 864.21 & 1.63  & 1.00  & \textbf{850.31} & \textbf{0.00}  & 2.24  & \underline{\textbf{811.08}} & \textbf{-4.61} & 2.95 \\
          & \multicolumn{1}{c}{75} & 1075.12 & \textbf{1075.12} & \textbf{0.00}  & 5.30  & 1123.61 & 4.51  & 11.61 & 1109.53 & 3.20  & 15.39 \\
          & \multicolumn{1}{c}{100} & 1199.09 & 1202.27 & 0.27  & 17.00 & \textbf{1199.09} & \textbf{0.00}  & 38.22 & \underline{\textbf{1184.76}} & \textbf{-1.20} & 43.49 \\
          & \multicolumn{1}{c}{175} & 1611.13 & \textbf{1611.13} & \textbf{0.00}  & 149.70 & 1659.37 & 2.99  & 42.31 & \underline{\textbf{1589.44}} & \textbf{-1.35} & 50.39 \\
          & \multicolumn{1}{c}{250} & 1862.44 & 1934.80 & 3.89  & 585.70 & \textbf{1862.44} & \textbf{0.00}  & 193.17 & \underline{\textbf{1852.54}} & \textbf{-0.53} & 249.39 \\
          & Average &       &       & 0.83  & 108.39 &       & 1.68  & 41.24 &       & -0.75 & 51.91 \\
          \midrule
    \multicolumn{1}{c}{\multirow{8}[0]{*}{\begin{sideways}$\alpha$ = 3\end{sideways}}}
          & \multicolumn{1}{c}{10} & 435.38 & \textbf{435.38} & \textbf{0.00}  & 0.00  & 439.60 & 0.97  & 0.13  & 435.53 & 0.03  & 0.93 \\
          & \multicolumn{1}{c}{20} & 567.49 & \textbf{567.49} & \textbf{0.00}  & 0.00  & 586.74 & 3.39  & 0.99  & 573.94 & 1.14  & 1.44 \\
          & \multicolumn{1}{c}{50} & 746.93 & 760.25 & 1.78  & 1.20  & \textbf{746.93} & \textbf{0.00}  & 2.14  & \underline{\textbf{715.23}} & \textbf{-4.24} & 2.90 \\
          & \multicolumn{1}{c}{75} & 929.58 & \textbf{929.58} & \textbf{0.00}  & 7.40  & 945.34 & 1.70  & 11.22 & 931.13 & 0.17  & 14.30 \\
          & \multicolumn{1}{c}{100} & 1058.80 & \textbf{1058.80} & \textbf{0.00}  & 23.10 & 1202.64 & 13.59 & 37.86 & \underline{\textbf{1051.00}} & \textbf{-0.74} & 40.43 \\
          & \multicolumn{1}{c}{175} & 1432.96 & \textbf{1432.96} & \textbf{0.00}  & 212.90 & 1469.92 & 2.58  & 41.58 & 1439.44 & 0.45  & 48.33 \\
          & \multicolumn{1}{c}{250} & 1635.12 & 1706.79 & 4.38  & 862.00 & \textbf{1635.12} & \textbf{0.00}  & 196.27 & 1710.56 & 4.61  & 285.03 \\
          & Average &       &       & 0.88  & 158.09 &       & 3.17  & 41.46 &       & 0.20  & 56.19 \\
          \bottomrule
    \end{tabular}%
    \end{adjustbox}
  \label{table:doublecenter}%
\end{table}%

\subsubsection{Results for Freitas and Penna (2020) benchmark set} \label{sec:tsplib}

\begin{table}[!ht]
  \footnotesize
  \centering
  \caption{HTGVNS results for \cite{FreitasPenna2020} instances}
  \begin{adjustbox}{max width=\textwidth}
    \begin{tabular}{lrrrrrrrrrrr}
    \toprule
          &       & \multicolumn{3}{c}{$HGVNS$} & \multicolumn{6}{c}{$HTGVNS$} \\

    \cmidrule(l){3-5} \cmidrule(l){6-11}
    \multicolumn{1}{c}{Instance} & \multicolumn{1}{c}{n} & \multicolumn{1}{c}{Sol.} & \multicolumn{1}{c}{$\overline{Sol.}$} & \multicolumn{1}{c}{Time} & \multicolumn{1}{c}{Sol.} & \multicolumn{1}{c}{Gap} & \multicolumn{1}{c}{$\overline{Sol.}$} & \multicolumn{1}{c}{$\overline{Gap}$} & \multicolumn{1}{c}{TSP Time} & \multicolumn{1}{c}{Time} \\
    \midrule
    berlin52 & 52    & 210.03 & 220.23 & 6.51  & \underline{\textbf{204.40}} & \underline{\textbf{-2.68}} & 209.31 & -0.33 & 0.01  & 4.20 \\
    bier127 & 127   & 3456.80 & 3587.88 & 53.74 & \underline{\textbf{3359.51}} & \underline{\textbf{-2.81}} & 3401.24 & -1.55 & 0.05  & 66.32 \\
    ch130 & 130   & 178.16 & 180.40 & 44.22 & \underline{\textbf{168.50}} & \underline{\textbf{-5.42}} & 190.38 & 6.78  & 0.09  & 59.03 \\
    d198  & 198   & 461.83 & 461.83 & 67.84 & \underline{\textbf{452.55}} & \underline{\textbf{-2.01}} & 469.83 & 1.73  & 0.15  & 51.33 \\
    eil51 & 51    & 13.45 & 13.68 & 11.59 & \underline{\textbf{11.74}} & \underline{\textbf{-12.74}} & 11.93 & -11.11 & 0.02  & 7.31 \\
    eil76 & 76    & 16.35 & 16.68 & 27.16 & \underline{\textbf{15.29}} & \underline{\textbf{-6.47}} & 16.11 & -1.44 & 0.02  & 30.93 \\
    kroA100 & 100   & 587.80 & 609.71 & 30.99 & \underline{\textbf{575.26}} & \underline{\textbf{-2.13}} & 591.32 & 0.58  & 0.04  & 38.94 \\
    kroA150 & 150   & 764.42 & 780.93 & 41.02 & \underline{\textbf{739.49}} & \underline{\textbf{-3.26}} & 758.44 & -0.77 & 0.07  & 38.33 \\
    kroA200 & 200   & 870.65 & 873.99 & 46.84 & \underline{\textbf{827.60}} & \underline{\textbf{-4.94}} & 869.43 & -0.14 & 0.31  & 44.39 \\
    kroB150 & 150   & 763.15 & 773.72 & 50.48 & \underline{\textbf{713.63}} & \underline{\textbf{-6.49}} & 744.35 & -2.43 & 0.28  & 43.39 \\
    kroB200 & 200   & 835.43 & 838.40 & 32.88 & \underline{\textbf{815.97}} & \underline{\textbf{-2.33}} & 831.43 & -0.48 & 0.94  & 63.40 \\
    kroC100 & 100   & 658.38 & 660.93 & 36.66 & \underline{\textbf{619.94}} & \underline{\textbf{-5.84}} & 639.42 & -2.87 & 0.03  & 48.12 \\
    kroD100 & 100   & 606.45 & 652.34 & 40.21 & \underline{\textbf{570.98}} & \underline{\textbf{-5.85}} & 593.43 & -2.00 & 0.06  & 54.80 \\
    kroE100 & 100   & 651.31 & 659.48 & 48.74 & \underline{\textbf{620.68}} & \underline{\textbf{-4.70}} & 639.32 & -1.82 & 0.17  & 60.13 \\
    lin105 & 105   & 378.25 & 380.43 & 40.33 & \underline{\textbf{377.46}} & \underline{\textbf{-0.21}} & 382.55 & 1.13  & 0.06  & 61.32 \\
    pr107 & 107   & 1204.42 & 1224.35 & 32.56 & \underline{\textbf{1128.71}} & \underline{\textbf{-6.29}} & 1198.64 & -0.47 & 0.02  & 52.40 \\
    pr124 & 124   & 1653.80 & 1996.62 & 25.67 & \underline{\textbf{1632.82}} & \underline{\textbf{-1.27}} & 1785.73 & 6.61  & 0.22  & 49.30 \\
    pr136 & 136   & 2642.00 & 2789.00 & 45.13 & \underline{\textbf{2595.17}} & \underline{\textbf{-1.77}} & 2639.32 & -0.10 & 0.63  & 52.39 \\
    pr144 & 144   & 1666.25 & 1675.75 & 43.42 & \underline{\textbf{1666.25}} & \underline{\textbf{0.00}} & 1707.47 & 2.46  & 0.09  & 48.20 \\
    pr152 & 152   & 2114.04 & 2128.53 & 61.51 & \underline{\textbf{2082.03}} & \underline{\textbf{-1.51}} & 2187.58 & 3.45  & 0.22  & 70.33 \\
    rat99 & 99    & 37.15 & 37.33 & 35.41 & \underline{\textbf{33.93}} & \underline{\textbf{-8.67}} & 37.65 & 1.34  & 0.04  & 39.87 \\
    rat195 & 195   & 71.40 & 71.93 & 46.78 & \underline{\textbf{67.17}} & \underline{\textbf{-5.92}} & 69.42 & -2.75 & 1.89  & 70.77 \\
    rd100 & 100   & 240.46 & 243.84 & 34.69 & \underline{\textbf{224.11}} & \underline{\textbf{-6.80}} & 236.42 & -1.66 & 0.82  & 43.59 \\
    st70  & 70    & 20.50 & 21.00 & 3.86  & \underline{\textbf{17.14}} & \underline{\textbf{-16.40}} & 22.95 & 11.67 & 0.01  & 10.83 \\
    \midrule
    Average &       &       & 37.58 &       & -4.86 & 0.26  & 46.23 \\
    \bottomrule
    \end{tabular}%
    \end{adjustbox}
  \label{tab:tsplib}%
\end{table}%

The latest FSTSP benchmark set used to test the HTGVNS performance is the one proposed by \cite{FreitasPenna2020}. The authors introduced 25 FSTSP instances based on the TSPLIB Symmetric Traveling Salesman Problem instances \citep{TSPLIB}.
These instances follow the FSTSP definition of \cite{Murray2015}. A different metric is considered to the vehicles travel distance. This is a reasonable consideration since drones are not affected by congestion, and they can fly in a straight line without considering the street network. On the other hand, trucks must respect traffic sign regulation and follow street network. The Euclidean metric is used to describe the drone travel distance, and the Manhattan metric is adopted to simulate the city block, thus, determining the truck distance.

The eligible drone customers are randomly generated such that for every instance there are between 80\% and 90\% serviceable customers. Service times $s^L$ and $s^R$ are unitary.
Moreover, both vehicles speed is 40 km/h, and the drone endurance is 40 minutes.

Table~\ref{tab:tsplib} provides the results obtained by HTGVNS. Note that column TSP Time indicates the runtime required to obtain the optimal TSP solution by Concorde.
These results consider ten HTGVNS executions for each instance available online\footnote{The instances and their solutions are available at \url{http://goal.ufop.br/fstsp/}}. 
Service time causes a large impact in instances with short travel times because the truck may need to wait a long time for the drone. 
Moreover, instances with customers disposed along a road are less impacted by the drone presence, since customers would be on the truck's way and it would be natural for the truck to visit them.
This behavior occurs in instance pr144 which prevents a significant improvement.





\section{Conclusions} \label{sec:Conclusions}

The Flying Sidekick Traveling Salesman Problem (FSTSP) concerns a Hamiltonian Cycle Problem \citep{Karp1972} generalization which poses interesting characteristics of a new modality of parcel distribution. 
The FSTSP seeks to coordinate a traditional delivery truck with a drone that may be launched from the truck. 
The problem considers different vehicle speeds and drone endurance to minimize the required time to serve all customers.

A Mixed Integer Programming formulation was proposed, solving multiple previously unsolved instances with up to 10 customers. 
In addition to providing new optimal solutions, it also provided stronger bounds than those obtained from formulations previously proposed for the FSTSP \citep{Murray2015, Ponza2016}.
However, due to the NP-hard characteristic of the problem, only small-scale instances could be solved. Therefore, a metaheuristic for finding good solutions for large-sized instances of the FSTSP was introduced, called Hybrid Tabu General Variable Neighborhood Search (HTGVNS). The HTGVNS algorithm initially constructs a solution by using a TSP solver to build the truck tour. This solution is then improved by the meta-heuristic General Variable Neighborhood Search (GVNS) mixed with a Tabu Search list to avoid cycling. 

The solution approaches were validated through numerical analysis that indicates the potential of the truck-drone delivery system to improve the operation.
The HTGVNS was evaluated considering three benchmark sets.
For the FSTSP sets, the algorithm improved all best-known solutions. On average, these solutions were improved by 3.25\%. For the TSP-D instances, the heuristic improved or achieved the same result for 1109 instances. 

Furthermore, the TSP-D, an FSTSP variant proposed by \cite{Agatz2018}, was also studied. The numerical analysis indicated that this delivery system might be more efficient by employing drones traveling faster than trucks. For this benchmark set, the best solution found consider instances in which the drone travels twice as fast as the truck, and the smallest improvements were observed for instances where both vehicles present the same speed.




\bibliographystyle{elsarticle-harv}
\section*{References}
\bibliography{mybibfile}

\begin{thebibliography}{38}
\expandafter\ifx\csname natexlab\endcsname\relax\def\natexlab#1{#1}\fi
\expandafter\ifx\csname url\endcsname\relax
  \def\url#1{\texttt{#1}}\fi
\expandafter\ifx\csname urlprefix\endcsname\relax\def\urlprefix{URL }\fi

\bibitem[{Agatz et~al.(2018)Agatz, Bouman, and Schmidt}]{Agatz2018}
Agatz, N., Bouman, P., Schmidt, M., 2018. Optimization approaches for the
  traveling salesman problem with drone. Transportation Science 52~(4),
  965--981.

\bibitem[{Applegate et~al.(2016)Applegate, Bixby, Chvatal, and Cook}]{Concorde}
Applegate, D., Bixby, R., Chvatal, V., Cook, W., 2016. Concorde tsp solver.
  Accessed: 07 June 2018.
\newline\urlprefix\url{http://www.math.uwaterloo.ca/tsp/concorde.html}

\bibitem[{Arenzana et~al.(2020)Arenzana, Macias, and Angeloudis}]{Arenzana2020}
Arenzana, A.~O., Macias, J. J.~E., Angeloudis, P., 2020. Design of hospital
  delivery networks using unmanned aerial vehicles. Transportation Research
  Record 2674~(5), 405--418.
\newline\urlprefix\url{https://doi.org/10.1177/0361198120915891}

\bibitem[{Beasley(1983)}]{Beasley1983}
Beasley, J., 1983. Route first—cluster second methods for vehicle routing.
  Omega 11~(4), 403--408.
\newline\urlprefix\url{https://www.sciencedirect.com/science/article/pii/0305048383900336}

\bibitem[{Bouman et~al.(2015)Bouman, Agatz, and Schmidt}]{Bouman2015}
Bouman, P., Agatz, N., Schmidt, M., 2015. Instances for the tsp with drone.

\bibitem[{Bouman et~al.(2018)Bouman, Agatz, and Schmidt}]{Boumanetal2017}
Bouman, P., Agatz, N., Schmidt, M., 2018. Dynamic programming approaches for
  the traveling salesman problem with drone. Networks 72~(4), 528--542.
\newline\urlprefix\url{https://onlinelibrary.wiley.com/doi/abs/10.1002/net.21864}

\bibitem[{Chase and Chu(2015)}]{Murray2015}
Chase, M.~C., Chu, A.~G., 2015. The flying sidekick traveling salesman problem:
  Optimization of drone-assisted parcel delivery. Transportation Research Part
  C: Emerging Technologies 54, 86--109.
\newline\urlprefix\url{https://www.sciencedirect.com/science/article/pii/S0968090X15000844}

\bibitem[{Chase and Ritwik(2020)}]{Murray2020}
Chase, M.~C., Ritwik, R., 2020. The multiple flying sidekicks traveling
  salesman problem: Parcel delivery with multiple drones. Transportation
  Research Part C: Emerging Technologies 110, 368--398.
\newline\urlprefix\url{https://www.sciencedirect.com/science/article/pii/S0968090X19302505}

\bibitem[{Croes(1958)}]{Croes1958}
Croes, G.~A., 1958. A method for solving traveling-salesman problems.
  Operations Research 6~(6), 791--812.
\newline\urlprefix\url{http://www.jstor.org/stable/167074}

\bibitem[{Dayarian et~al.(2020)Dayarian, Savelsbergh, and
  Clarke}]{Dayarian2020}
Dayarian, I., Savelsbergh, M., Clarke, J.-P., 2020. Same-day delivery with
  drone resupply. Transportation Science 54~(1), 229--249.
\newline\urlprefix\url{https://doi.org/10.1287/trsc.2019.0944}

\bibitem[{de~Freitas and Penna(2020)}]{FreitasPenna2020}
de~Freitas, J.~C., Penna, P. H.~V., 2020. A variable neighborhood search for
  flying sidekick traveling salesman problem. International Transactions in
  Operational Research 27~(1), 267--290.

\bibitem[{Dell'Amico et~al.(2020)Dell'Amico, Montemanni, and
  Novellani}]{DellAmico2020}
Dell'Amico, M., Montemanni, R., Novellani, S., 2020. Matheuristic algorithms
  for the parallel drone scheduling traveling salesman problem. Annals of
  Operations Research 289, 211--226.
\newline\urlprefix\url{https://link.springer.com/article/10.1007/s10479-020-03562-3}

\bibitem[{Dorling et~al.(2017)Dorling, Heinrichs, Messier, and
  Magierowski}]{Dorlingetal2018}
Dorling, K., Heinrichs, J., Messier, G.~G., Magierowski, S., 2017. Vehicle
  routing problems for drone delivery. IEEE Transactions on Systems, Man, and
  Cybernetics: Systems 47~(1), 70--85.

\bibitem[{Douglas(2020)}]{Douglas2020}
Douglas, A., apr 2020. Flytrex takes-off with drone deliveries following
  covid-19 social distancing protocols.
\newline\urlprefix\url{https://bit.ly/3dzQD48}

\bibitem[{Ferrandez et~al.(2016)Ferrandez, Harbison, Weber, Sturges, and
  Rich}]{Ferrandezetal-2016}
Ferrandez, S., Harbison, T., Weber, T., Sturges, R., Rich, R., 04 2016.
  Optimization of a truck-drone in tandem delivery network using k-means and
  genetic algorithm. Journal of Industrial Engineering and Management 9, 374.

\bibitem[{Glover and Laguna(1998)}]{Glover1999}
Glover, F., Laguna, M., 1998. Tabu Search. Vol. 1--3. Springer US, pp.
  2093--2229.
\newline\urlprefix\url{https://doi.org/10.1007/978-1-4613-0303-9_33}

\bibitem[{Gonzalez-R et~al.(2020)Gonzalez-R, Canca, Andrade-Pineda, Calle, and
  Leon-Blanco}]{Gonzalezetal2020}
Gonzalez-R, P.~L., Canca, D., Andrade-Pineda, J.~L., Calle, M., Leon-Blanco,
  J.~M., 2020. Truck-drone team logistics: A heuristic approach to multi-drop
  route planning. Transportation Research Part C: Emerging Technologies 114,
  657--680.
\newline\urlprefix\url{https://www.sciencedirect.com/science/article/pii/S0968090X19309568}

\bibitem[{Guillot(2020)}]{Guillot2020}
Guillot, G., apr 2020. {Using Drones for Food Deliveries, Social Distancing,
  and Health Monitoring: Drones in America MarketScale}.
\newline\urlprefix\url{https://bit.ly/2LnyzOO}

\bibitem[{Ha et~al.(2018)Ha, Deville, Pham, and Hà}]{Ha2018}
Ha, Q.~M., Deville, Y., Pham, Q.~D., Hà, M.~H., 2018. On the min-cost
  traveling salesman problem with drone. Transportation Research Part C:
  Emerging Technologies 86, 597--621.
\newline\urlprefix\url{https://www.sciencedirect.com/science/article/pii/S0968090X17303327}

\bibitem[{Hansen et~al.(2010)Hansen, Mladenovi{\'{c}}, and Pérez}]{Hansen2008}
Hansen, P., Mladenovi{\'{c}}, N., Pérez, J. A.~M., 02 2010. Variable
  neighbourhood search: Methods and applications. 4OR 175, 367--407.

\bibitem[{Hansen et~al.(2016)Hansen, Mladenovic, Todosijević, and
  Hanafi}]{Hansen2017}
Hansen, P., Mladenovic, N., Todosijević, R., Hanafi, S., 08 2016. Variable
  neighborhood search: basics and variants. EURO Journal on Computational
  Optimization 5.

\bibitem[{Jeong et~al.(2019)Jeong, Song, and Lee}]{Jeong2019}
Jeong, H.~Y., Song, B.~D., Lee, S., 2019. Truck-drone hybrid delivery routing:
  Payload-energy dependency and no-fly zones. International Journal of
  Production Economics 214, 220--233.
\newline\urlprefix\url{https://www.sciencedirect.com/science/article/pii/S0925527319300106}

\bibitem[{Karak and Abdelghany(2019)}]{Karak2019}
Karak, A., Abdelghany, K., 2019. The hybrid vehicle-drone routing problem for
  pick-up and delivery services. Transportation Research Part C: Emerging
  Technologies 102, 427--449.
\newline\urlprefix\url{https://www.sciencedirect.com/science/article/pii/S0968090X18312932}

\bibitem[{Karp(1972)}]{Karp1972}
Karp, R.~M., 1972. Reducibility among Combinatorial Problems. Springer US, pp.
  85--103.

\bibitem[{Mercer(2018)}]{Drone_Applications2018}
Mercer, C., May 2018. How are drones used? top companies using drones right
  now.
  \url{https://www.techworld.com/picture-gallery/apps-wearables/best-uses-of-drones-3605145/},
  accessed 30 May 2018.

\bibitem[{Moshref-Javadi et~al.(2020)Moshref-Javadi, Hemmati, and
  Winkenbach}]{MoshrefJavadi2020}
Moshref-Javadi, M., Hemmati, A., Winkenbach, M., 2020. A truck and drones model
  for last-mile delivery: A mathematical model and heuristic approach. Applied
  Mathematical Modelling 80, 290--318.
\newline\urlprefix\url{https://www.sciencedirect.com/science/article/pii/S0307904X19306936}

\bibitem[{Otto et~al.(2018)Otto, Agatz, Campbell, Golden, and Pesch}]{Otto2018}
Otto, A., Agatz, N., Campbell, J., Golden, B., Pesch, E., 2018. Optimization
  approaches for civil applications of unmanned aerial vehicles (uavs) or
  aerial drones: A survey. Networks 72~(4), 411--458.
\newline\urlprefix\url{https://onlinelibrary.wiley.com/doi/abs/10.1002/net.21818}

\bibitem[{Penna et~al.(2013)Penna, Subramanian, and Ochi}]{Penna2013}
Penna, P. H.~V., Subramanian, A., Ochi, L.~S., 2013. {An iterated local search
  heuristic for the heterogeneous fleet vehicle routing problem}. Journal of
  Heuristics 19~(2), 201--232.

\bibitem[{Poikonen et~al.(2017)Poikonen, Wang, and Golden}]{Poikonen2017}
Poikonen, S., Wang, X., Golden, B., 2017. The vehicle routing problem with
  drones: Extended models and connections. Networks 70~(1), 34--43.
\newline\urlprefix\url{https://onlinelibrary.wiley.com/doi/abs/10.1002/net.21746}

\bibitem[{Ponza(2016)}]{Ponza2016}
Ponza, A., 04 2016. Optimization of drone-assisted parcel delivery. Ph.D.
  thesis, Universit a Degli Studi Di Padova.

\bibitem[{Pugliese and Guerriero(2017)}]{DiPugliaPugliese2017}
Pugliese, L. D.~P., Guerriero, F., 2017. Last-mile deliveries by using drones
  and classical vehicles. In: Sforza, A., Sterle, C. (Eds.), Optimization and
  Decision Science: Methodologies and Applications. Springer International
  Publishing, pp. 557--565.

\bibitem[{Saleu et~al.(2018)Saleu, Deroussi, Feillet, Grangeon, and
  Quilliot}]{MbiadouSaleu2018}
Saleu, R. G.~M., Deroussi, L., Feillet, D., Grangeon, N., Quilliot, A., 2018.
  An iterative two-step heuristic for the parallel drone scheduling traveling
  salesman problem. Networks 72~(4), 459--474.
\newline\urlprefix\url{https://onlinelibrary.wiley.com/doi/abs/10.1002/net.21846}

\bibitem[{Schermer et~al.(2019)Schermer, Moeini, and Wendt}]{Schermer2019}
Schermer, D., Moeini, M., Wendt, O., 2019. A matheuristic for the vehicle
  routing problem with drones and its variants. Transportation Research Part C:
  Emerging Technologies 106, 166--204.
\newline\urlprefix\url{https://www.sciencedirect.com/science/article/pii/S0968090X19301081}

\bibitem[{Skorobohatyj(1995)}]{TSPLIB}
Skorobohatyj, G., 1995. {TSPLIB}.
\newline\urlprefix\url{https://www.iwr.uni-heidelberg.de/groups/comopt/}

\bibitem[{Song et~al.(2018)Song, Park, and Kim}]{Song2018}
Song, B.~D., Park, K., Kim, J., 2018. Persistent uav delivery logistics: Milp
  formulation and efficient heuristic. Computers and Industrial Engineering
  120, 418--428.
\newline\urlprefix\url{https://www.sciencedirect.com/science/article/pii/S0360835218302146}

\bibitem[{Souza et~al.(2010)Souza, Coelho, Ribas, Santos, and
  Merschmann}]{Souza2010}
Souza, M. J.~F., Coelho, I.~M., Ribas, S., Santos, H.~G., Merschmann, L. H.~C.,
  2010. A hybrid heuristic algorithm for the open-pit-mining operational
  planning problem. European Journal of Operational Research 207~(2),
  1041--1051.

\bibitem[{Ulmer and Thomas(2018)}]{Ulmer2017}
Ulmer, M.~W., Thomas, B.~W., 2018. Same-day delivery with heterogeneous fleets
  of drones and vehicles. Networks 72~(4), 475--505.
\newline\urlprefix\url{https://onlinelibrary.wiley.com/doi/abs/10.1002/net.21855}

\bibitem[{Wang et~al.(2017)Wang, Poikonen, and Golden}]{Wang2017}
Wang, X., Poikonen, S., Golden, B., 04 2017. The vehicle routing problem with
  drones: Several worst-case results. Optimization Letters 11.

\end{thebibliography}

\end{document}